\documentclass[fleqn,12pt]{olplainarticle}
\usepackage[T1]{fontenc}
\usepackage{helvet}

\usepackage{graphicx}
\usepackage{xcolor}
\usepackage[labelfont=bf,font=sf]{caption} 
\usepackage{float} 
\usepackage{comment}
\usepackage{amsmath}
\usepackage{xcolor}
\usepackage{titlesec}
\usepackage{setspace}
\usepackage{pdfpages}

\title{Fin ray-inspired, Origami, Small Scale Actuator for Fin Manipulation in Aquatic Bioinspired Robots}

\author{

Minh Vu$^{1,2\ast}$, Revathy Ravuri$^{1}$, Angus Muir$^{1}$, Charles Mackie$^{1}$, Andrew Weightman$^{3,4}$, Simon Watson$^{1,4}$, Tim J. Echtermeyer$^{1,2,5\ast}$\\

\normalsize{$^{1}$Department of Electrical and Electronic Engineering, The University of Manchester}\\
\normalsize{$^{2}$Photon Science Institute, The University of Manchester}\\
\normalsize{$^{3}$School of Engineering, Faculty of Science and Engineering, The University of Manchester}\\
\normalsize{$^{4}$Centre for Robotics and Artificial Intelligence, The University of Manchester}\\
\normalsize{$^{5}$National Graphene Institute, The University of Manchester}\\
\normalsize{Manchester, M13 9PL, United Kingdom}\\
\normalsize{$^\ast$Correspondence:  minh.vu@manchester.ac.uk, tim.echtermeyer@manchester.ac.uk.}
}

\date{}


\begin{abstract}
\sffamily
Fish locomotion is enabled by fin rays---actively deformable boney rods, which manipulate the fin to facilitate complex interaction with surrounding water and enable propulsion. Replicating the performance and kinematics of the biological fin ray from an engineering perspective is a challenging task and has not been realised thus far. This work introduces a prototype of a fin ray-inspired origami electromagnetic tendon-driven (FOLD) actuator, designed to emulate the functional dynamics of fish fin rays. Constructed in minutes using origami/kirigami and paper joinery techniques from flat laser-cut polypropylene film, this actuator is low-cost at £0.80 (\$1), simple to assemble, and durable for over one million cycles. We leverage its small size to embed eight into two fin membranes of a 135~mm long cuttlefish robot capable of four degrees of freedom swimming. We present an extensive kinematic and swimming parametric study with 1015 data points from 7.6 hours of video, which has been used to determine optimal kinematic parameters and validate theoretical constants observed in aquatic animals. Notably, the study explores the nuanced interplay between undulation patterns, power distribution, and locomotion efficiency, underscoring the potential of the actuator as a model system for the investigation of energy-efficient propulsion and control of bioinspired systems. The versatility of the actuator is further demonstrated by its integration into a fish and a jellyfish.

\end{abstract}

\begin{document} 
\flushbottom
\maketitle 
\setstretch{1.25}



\section*{Introduction}
Fish are extraordinary swimmers. Their ability to navigate and manoeuvre underwater with precision and efficiency has long fascinated scientists and engineers seeking to further understand the hydrodynamic interaction\cite{lauder_hydrodynamics_2005, triantafyllou_hydrodynamics_2000} and develop novel aquatic propulsion systems \cite{low_biomimetic_2007,youssef_design_2022}. Central to their remarkable locomotion is the intricate movement and control of their fins, which act as multi-directional control surfaces that can efficiently manipulate the surrounding water \cite{bale_separability_2014,godoy-diana_diverse_2018,fish_passive_2006,smits_undulatory_2019}. In particular, the fin rays (\emph{lepidotrichia}) in ray-finned fishes (\emph{Actinopterygii}) demonstrate a high level of individual morphing ability \cite{hannard_segmentations_2021,alben_mechanics_2007,videler_relation_1986}, which collectively allows for three-dimensional deformation of the fin. Replicating this elaborate functionality in a compact and easily manufacturable actuator poses a significant engineering challenge, requiring a combination of innovative design, suitable material properties, advanced manufacturing techniques, and precise control mechanisms. Overcoming these challenges would enable the creation of versatile robotic systems that mimic biological fluid dynamics, opening up possibilities for environmentally adaptive robots capable of complex, autonomous underwater tasks and interactions.



\begin{figure}[htbp!]
    \centering
    \includegraphics[width=1\textwidth]{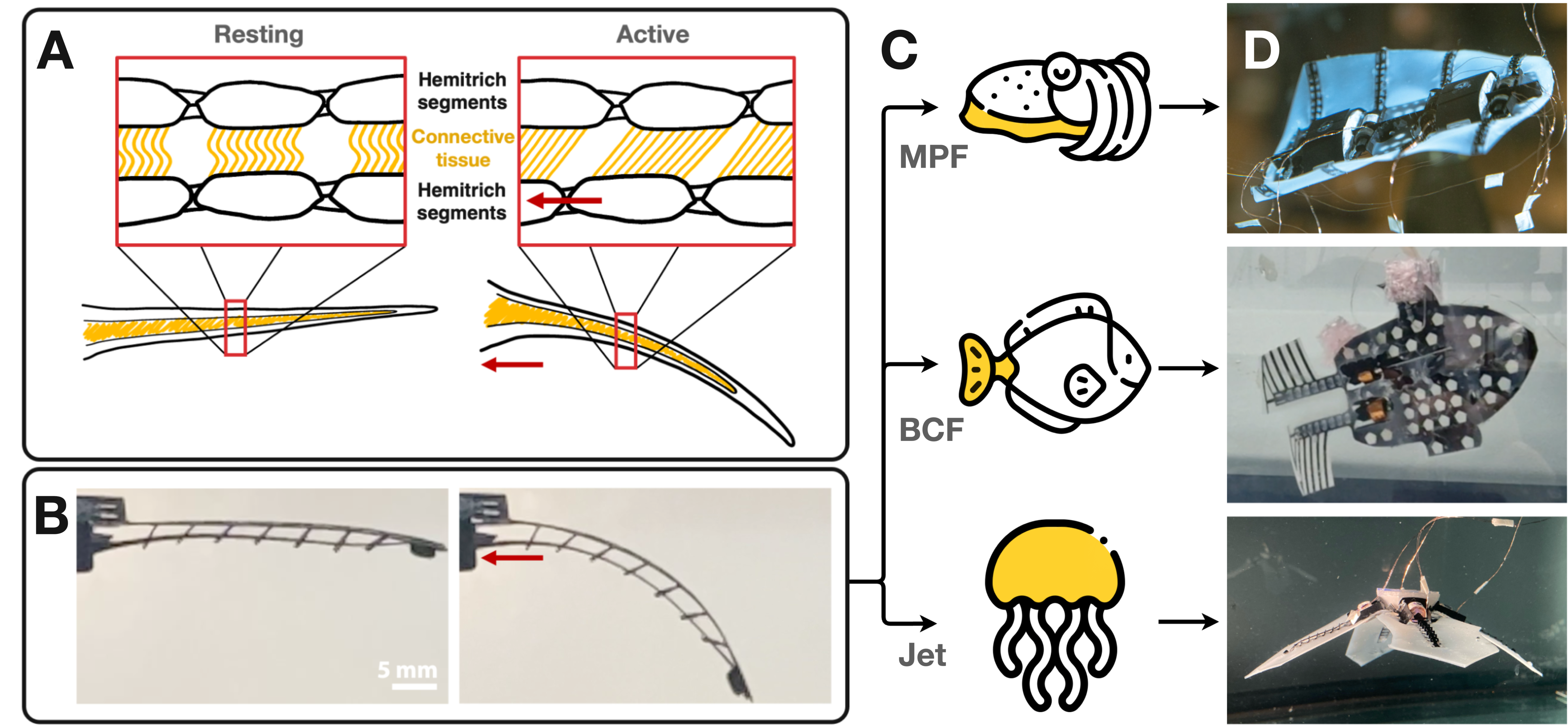}
    \caption{\textbf{Fin ray-inspired actuator for aquatic robots} \\{\color{gray}\small \textbf{(A)} A fin ray \textit{(lepidotrichium)} at rest and while bending. The yellow parts represents the collagen fibrils. The red arrow represents the pulling force \textbf{(B)} FOLD actuator mimicking bending in fin ray. Pulling the tendon causes deflection in the fin ray actuator. \textbf{(C)} Aquatic locomotion, the yellow highlight indicates the propulsion mechanism. From top to bottom: Median paired fin (MPF) propulsion, body-caudal fin (BCF) propulsion, jet propulsion. \textbf{(D)} FOLD actuator integrated in three distinct bioinspired robot morphologies: cuttlefish, fish, jellyfish.}}
    \label{fig:introduction}
\end{figure}

Among ray-finned fishes, the fin ray serves two important purposes---stiffening the fin under hydrodynamic load and bending the fin to interact with the water, utilised 
to achieve complex fin movements and propulsion. The fin ray is composed of two parallel halves (\emph{hemitrich}); convex segmented flexible beams that are tapered and joined at their tips\cite{videler_relation_1986,hannard_segmentations_2021,alben_mechanics_2007}. They are linked together along the length by a compliant core, which also connects them to the fin's membrane (Fig.~\ref{fig:introduction}A). Groups of antagonistic muscles are attached to the base of each hemitrich; as they contract, one hemitrich is displaced relative to the other, creating a shear moment, resulting in flexural deformation of the fin ray \cite{hannard_segmentations_2021,videler_relation_1986,lauder_design_2005,alben_mechanics_2007}. The complex interaction between the segmented hemitrichs and the musculature generates a finely tuned mechanical response, allowing the fins to achieve intricate deformation\cite{hannard_segmentations_2021,lauder_design_2005,nguyen_curvature-induced_2017}. The amalgamation of these individual deformations results in the emergence of three-dimensional shapes from the two-dimensional fin membrane. While much research has concentrated on the hydrodynamic advantages derived from the passive curvature and flexibility (uniform and non-uniform) of fins\cite{he_effects_2022,zhu_effects_2017,liu_fin_2017}, the areas of active curvature, tunable stiffness, and controlled deformation remain underexplored\cite{fernandez-gutierrez_effect_2020,nose_design_2021,quinn_tunable_2021}.

The pursuit of emulating the biomechanical principles and characteristics of fish fins has been driven by the increasing need for minimally invasive automation in aquaculture \cite{ubina_review_2022,zhang_farmed_2024}, marine monitoring \cite{katzschmann_exploration_2018,shree_design_2013}, and the study of efficient mode of propulsion \cite{fish_advantages_2020,daniel_bioinspired_2020,li_underwater_2023}. This breath of need has given rise to a diverse range of approaches. Notable work such as SoFi \cite{katzschmann_exploration_2018}, TunaBot Flex \cite{white_tunabot_2021}, dolphin robots \cite{chen_development_2022,yu_development_2016}, FinBot \cite{berlinger_fish-like_2021} chose to simplify the fin to a homogeneous water foil and instead focus on the musculature to mimic the movement seen in aquatic creatures such as tuna, salmon, and shark employing body-caudal-fin (BCF) locomotion (Fig. \ref{fig:introduction}C). Others, placing more emphasis on fin kinematics, used motor-connected rods embedded in a membrane to replicate the articulation of the fin, a technique often seen in robots using median-paired-fins (MPF) locomotion, inspired by the undulating fins of rays, knifefish, and cuttlefish\cite{xing_asymmetrical_2022,low_biomimetic_2006,zhu_effects_2017,chen_design_2022,yin_kinetic_2021,bianchi_bioinspired_2023}. Despite these advancements, achieving the nuanced fin control observed in nature remains elusive, relying on the passive curvature resulting from the flexibility of the materials used to construct the fin. Approaches using novel actuators like dielectric elastomer actuators, shape memory alloys, ionic polymer–metal composites have yielded varying results. Due to their inherent characteristics, the bending motion is naturally achieved and closely mirrors the dynamic bending of fin rays \cite{chen_novel_2011,berlinger_modular_2018,kim_design_2017,zhang_underwater_2023,shintake_soft_2018,chen_modeling_2010,li_fast-moving_2017,li_self-powered_2021}. However, they present their own set of challenges in power and control, which limits the configuration and number of actuators that can be implemented, ultimately constraining the design and performance of the robot. This dichotomy between complexity in function and simplicity in design poses a substantial challenge in developing dexterous bioinspired aquatic robots. The high entry cost of equipment, materials, and expertise creates a significant barrier for researchers, a problem that becomes even more pronounced as the size of the robot decreases.


We seek to unite these seemingly contradictory criteria by presenting the initial proof of concept design and characterisation of fin ray-inspired origami electromagnetic tendon-driven (FOLD) actuators (Fig. \ref{fig:introduction}B and Movie S1). The design is small, allowing for dense placement of actuators to enable dexterous manipulation of a bionic fin while remaining adaptable in various robotic systems. 
Constructed using origami/kirigami and paper
joinery techniques from a flat laser cut polypropylene film, the actuator is easily and cheaply fabricated (£0.80/\$1.00), without specialised equipment or materials. These design principles are further extended to holistically design three distinct underwater robot prototypes with a monolithic body with actuators: a cuttlefish robot, a tuna robot, and a jellyfish robot (Fig. \ref{fig:introduction}D), demonstrating the efficacy and versatility of the actuator. Among these, the cuttlefish robot demonstrates unprecedented capabilities in an aquatic robot of that size, performing various swimming modes to achieve four independent degrees of motion. Using videos, we were able to collect a large dataset to identify the optimal kinematic parameters and confirm theoretical numerical constants (such as Strouhal number \cite{taylor_flying_2003,eloy_optimal_2012} and Specific Wavelength \cite{bale_convergent_2015,nangia_optimal_2017}), as well as to investigate the power-thrust relationship. Moreover, facilitated by independent control over each fin ray, we explored the effects of wave envelope on propulsion\cite{zhang_computational_2007,he_development_2015}. These findings indicate that the platform holds tremendous potential for enhancing biomimetic locomotion models and studying complex hydrodynamic interactions.





\section*{RESULTS \& DISCUSSION}
\subsection*{Working principle and design overview}


We propose a flat-constructed push-pull tendon-driven actuator to emulate the antagonistic bending of fish fin rays. Measuring 33 mm long, 6 mm wide, and consisting of a backbone (Fig. \ref{fig:singleactuator}A) and tendon (Fig. \ref{fig:singleactuator}B) made from laser-cut polypropylene (PP) joined at the tip represent the structure of the hemitrichs. The tendon transmits tension and compression force. Any buckling of the tendon is limited by the series of hooks along the backbone, which act like the collagenous core in a fin ray to maintain the separation between the two halves while still allowing sliding motion. The bidirectional linear force is generated using a voice coil actuator (VCA), utilising Lorentz force, the magnetic field generated from a current running through a moving solenoid repels/attracts a permanent magnet. Depending on the direction of the current, the VCA pushes and pulls on the tendon, deflecting the fin ray. The design does not feature any sealed chambers, allowing water to flow around the coil and magnets. This allows for the coil to be water-cooled and the VCA to be water-lubricated. This design further minimises the need for any additional components and makes the actuator inherently pressure-tolerant.

\subsection*{Laser-cut fabrication}
We found that common polypropylene film folders with a thickness of 0.2 mm match the required range of flexibility and durability. Laser cutting allows for an affordable and quick method of iterative prototyping. Two different settings are used during laser-cutting: a cut mode with approximately 0.6 W laser power to fully cut through the PP sheet and an engraving mode with a reduced 0.18 W laser power to only partially cut the PP sheet and facilitate folds. These cuts and folds then allow the application of Origami/Kirigami, pop-up and paper joinery design principles to construct out-of-plane elements that can be mechanically joined without the use of adhesives\cite{whitney_pop-up_2011}. The total material cost of the actuator is less than £0.80 (\$1), the full breakdown is shown in Table S.1. 

\subsection*{Origami/Kirigami assembly}
\begin{figure}[htbp!]
    \centering
    \includegraphics[width=1\textwidth]{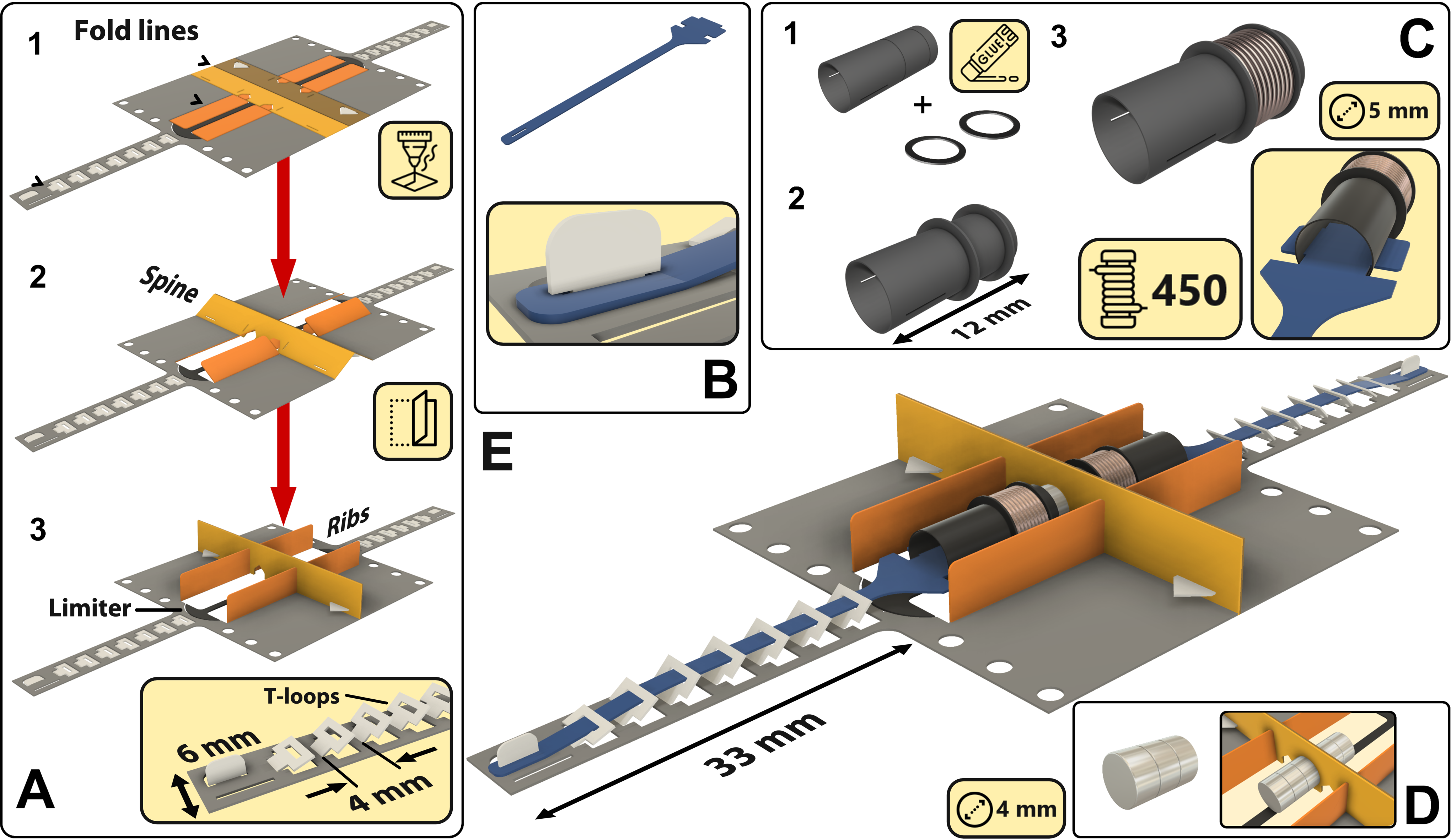}
    \caption{\textbf{Assembly of paired fin ray.} \\{\color{gray}\small(\textbf{A}) Folding assembly: 1. The polypropylene sheet is laser cut and engraved to create the desired profile; 2. The flat laser cut profile is folded to create the spine and ribs; 3. Tabs and slits are used to secure the structure. Tendon hooks are folded up. (\textbf{B}) The tendon: Laser-cut from the PP sheet and attached to the tab at the end of the backbone. (\textbf{C}) The assembly process of the coil: 1. The straw is laser cut to size; 2. The rings are glued on to create the coil bobbin; 3. Copper windings are added. The tendon is attached using the slits. (\textbf{D}) Magnets are attached on either side of the spine. (\textbf{E}) The coil is slid over the limiter (black) and magnets to complete the assembly.}}
    \label{fig:singleactuator}
\end{figure}

The assembly starts with step A (Figure \ref{fig:singleactuator}A), the support structure is folded into a T-shape, creating a spine that introduces rigidity along the length. Subsequently, the rib-shaped flaps are folded up 90\textdegree{} and attached to the spine with a clip mechanism, adding span-wise rigidity. The beam features two main elements: T-shaped loops and a tab at the end of the beam to which a tendon will be secured. The tab is in the shape of a sideways D, which folds up 90\textdegree{} to insert into the slit on the tendon. The T-loops have rectangular horizontal slits, measuring 1-by-2~mm, at the horizontal section of the T. They fold up to create evenly distributed standing loops along the backbone (inset Fig. \ref{fig:singleactuator}A). These T-loops act as guides for the tendon, suspending it 2~mm above the backbone and ensuring that the tendon and backbone will remain parallel as the fin ray flexes. Similar to the segmentation in a hemitrich, the cut-outs in the T-loops contribute to the flexibility of the backbone and ultimately the actuator's bending profile. In step B, laser cut tendons (200 µm thick PP) are slotted through the T-loops and secured to tab at the distal end of the backbone. A slit in the tendon slightly smaller than the width of the D-shape facilitates latching of the tendon with the D-shaped tab, as shown in Fig. \ref{fig:singleactuator}B. 

The VCA serves as the ``muscle" to flex the fin ray. Unlike the biological counterpart, which has antagonistic pairs of muscles to flex the fin ray, the voice coil actuator performs both push and pull motion. As such, a single VCA and a single tendon, located only on one side of the backbone, is sufficient to induce upward and downward bending, further simplifying the implementation of the fin ray. The stiffness of the tendon, originating from the PP sheet and its geometry, in combination with the tightly spaced T-hooks that limit buckling similar to the collagenous core in a fin ray, allows the application of both tensile and compressive forces to the tendon. The coil bobbin is manufactured using a 5~mm diameter polypropylene straw, which has a smooth low friction surface finish (Fig. \ref{fig:singleactuator}C). Slits are cut into the straw to attach the tendon, and two lines are engraved 4 mm apart to attach two 9 mm diameter rings, which secure the windings. The position of the windings on the bobbin, relative to static magnets assembled in the subsequent step, has been carefully chosen to ensure operating the VCA at its maximum mechanical output power point and with maximum travel range. More details on the manufacturing and characterisation of the VCA can be found in Supplementary Materials. 100 µm diameter polyimide-coated copper wire was used to wind the coil with the maximum possible windings of $\approx$ 450, resulting in a resistance of $\approx$ 25 $\Omega$ and inductance of $\approx$ 1~mH. The resulting coil measures 12.5~mm in length with an outer diameter of 9~mm.

In the final step, three circular 4 mm diameter Neodymium magnets of 2 mm thickness are attached to each side of the spine (Fig. \ref{fig:singleactuator}D). Here, the actuators are configured in a back-to-back arrangement of fin rays, i.e. symmetry of the design allows securing the magnets to the spine by magnetic forces between adjacent magnets on the left- and right-hand side of the spine. To prevent the coil from sliding off and away from the magnet, a limiter extends from underneath the magnets to limit the actuation stroke to ±2 mm. A semicircular arrowhead allows the coil to slide over during the assembly but prevents the coil from sliding back out. The VCA is capable of producing a peak force of 490 mN at a current of 480 mA, the force profile follows a bell curve shape along the stroke distance \cite{paul_design_2019}. Full characterisation of the force-displacement curves is shown in Supplementary Materials.

\subsection*{Deflection characterisation}



To characterise the deflection achieved by the FOLD actuator, we investigated three main relationships: current-deflection, transient response, and frequency response. Understanding these parameters is crucial for optimising the actuator's performance and limitations, especially under varying loads and dynamic conditions. By comparing the deflection behaviours when unloaded, loaded with weights at the tips, and embedded in a membrane, we aim to simulate realistic operational scenarios. All experiments are conducted underwater to replicate the intended operational context and to provide relevant data for potential aquatic applications.






To characterise the deflection profile of the fin ray actuator in water, the distal tip was tracked by processing videos of the motion at 240 frames per second (see Supplementary Materials). Figure \ref{fig:singleactuatorchracterisation}A illustrates the transition from maximum push to pull deflection at an excitation frequency of 0.5 Hz. At this low frequency, the fin ray achieves a maximum bending angle close to 90\textdegree{}, settling fully at the extremes and displaying a homogeneous bending profile with constant curvature along its length during the transition. At higher frequencies, such as 4 Hz, the deflection profile changes significantly, showing an asymmetric bending resembling a cupping motion due to the reactive mechanical force distribution along the fin ray (Movie S2). The tendon exerts force initially at the tip, with inertia causing the rest of the fin ray to lag, creating a cupping effect. This behaviour can potentially be exploited to increase the interaction of the fin ray with the surrounding water, as is commonly found in fish fin and has been associated with increased hydrodynamic performance\cite{blevins_rajiform_2012,lauder_design_2005,liu_fin_2017}.


\begin{figure}[htbp!]
    \centering
    \includegraphics[width=1\textwidth]{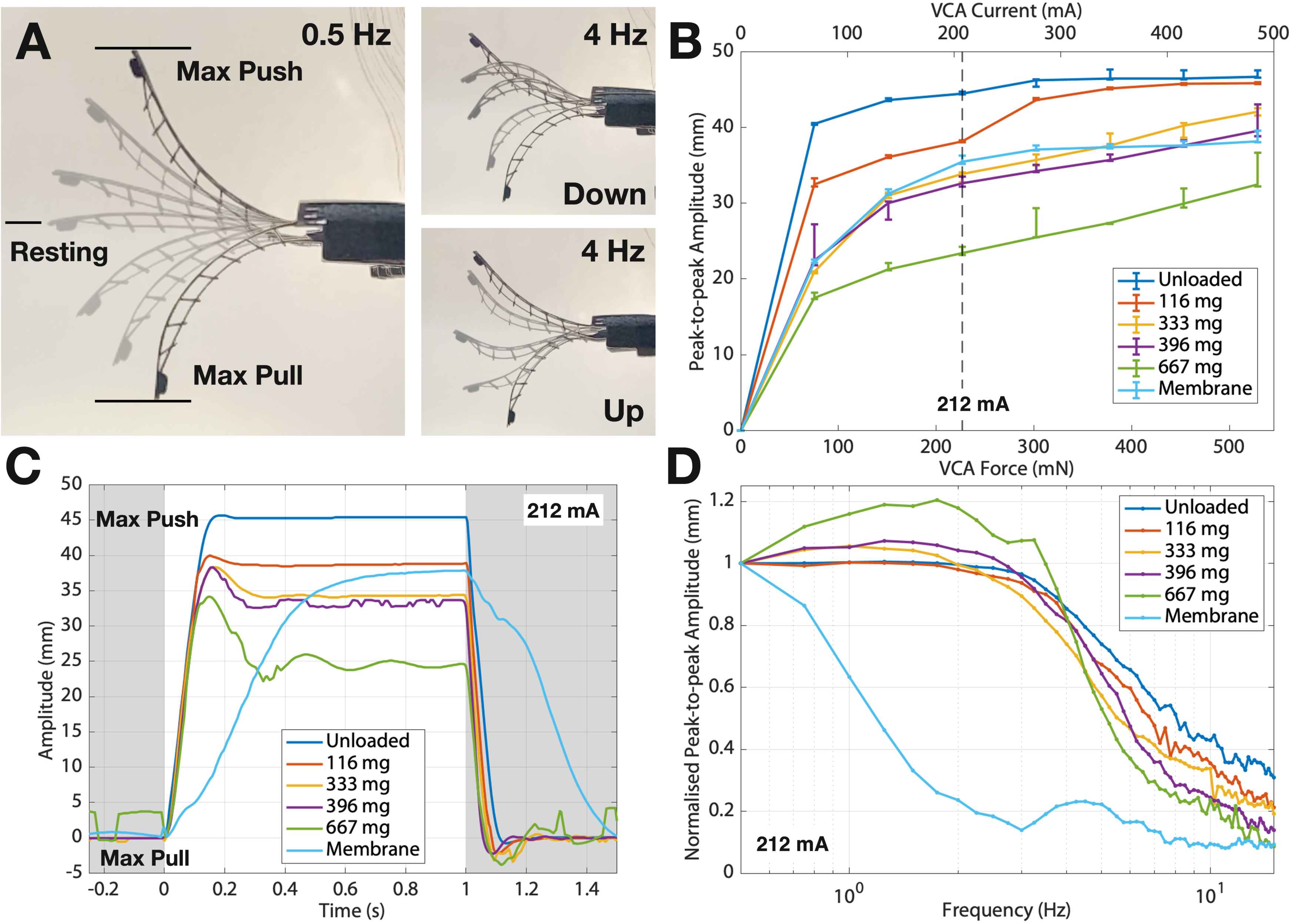}
    \caption{\textbf{Characterisation of the fin ray in water.} \\{\color{gray}\small(\textbf{A}) Left: Composite side-view stills of the fin ray oscillating at a frequency of 0.5 Hz. Right: Asymmetric curvature profile at a frequency of 4 Hz. (\textbf{B}) Peak-to-peak deflection amplitude as a function of the current of the fin ray loaded with different weights at the tip at a frequency of 0.5 Hz. The dashed line at a current of 212 mA marks the upper current limit before the actuator enters the saturation regime due to the mechanical limit.(\textbf{C}) Step response of the fin ray loaded with different weights, excited with a current step of 212 mA amplitude and 1 s duration (the white region indicates when the signal is on).(\textbf{D}) Frequency response of the fin ray under various load conditions (unloaded, loaded at the tip, enclosed in a membrane) under sinusoidal excitation with a peak current of 212 mA.}}
    \label{fig:singleactuatorchracterisation}
\end{figure}

Figure \ref{fig:singleactuatorchracterisation}B presents the peak-to-peak deflection of the fin ray under different loads as a function of driving current. The VCA was driven using a step signal, applying alternating negative and positive current steps with instantaneous transitions and varying hold times at the extremes to observe maximum push/pull deflection of the fin ray. The unloaded fin ray exhibits a maximum deflection of 47~mm, nearing to ±90\textdegree{} bending. This upper limit is imposed by the VCA's travel range of ±2 mm, corresponding to an 11.75:1 transmission ratio. Additional weights shift the neutral position towards the pull direction due to gravity and reduce the overall peak-to-peak deflection, maintaining an approximately linear deflection-current relationship. The majority of the deflection data show a small spread, highlighting the repeatability and controlability of the actuator, and simultaneously the robustness of the video capture setup. Currents above 212 mA lead to deflection saturation for both unloaded and lightly loaded (116 mg) fin rays, hence 212 mA was selected for subsequent characterisations to avoid saturation.


To determine the transient response of the fin ray, we captured the experiments on videos and used image processing to locate the tip. The fin ray, under different load conditions, was actively pulled to the maximum pull-deflection and then pushed with a current step of 212 mA of one second duration (Fig. \ref{fig:singleactuatorchracterisation}C). The weights of [116, 333, 396, 667] mg were achieved using combinations of plastic/metal screws and nuts (further details in Supplementary Materials). When unloaded, the fin ray exhibits a rise-time (10 to 90$\%$) of 100 ms, decreasing to 54 ms when a load of 667 mg is applied. However, this reduction in the rise time can be attributed to the diminished actuation range under load. The velocity of the tip is independent of the load, displaying constant velocities of 0.37 m/s.  When the beam is unloaded, a slight overshoot of 0.4 mm is observable, followed by rapid settling within 150 ms. Conversely, at the highest load of 667 mg, the overshoot becomes more pronounced at 9.5 mm with an increased settling time of 400 ms. 

As fish commonly employ sinusoidal movements for locomotion, characterising the frequency response of the fin ray under sinusoidal excitation is vital. The fin ray's frequency response was investigated by applying a sinusoidal input signal with frequencies ranging from 0.5 to 15 Hz in 0.25 Hz increments. The supply current was set at 212 mA, just below the saturation region, to ensure sinusoidal movement of the fin ray and prevent clipping of the actuation. As shown in Figure \ref{fig:singleactuatorchracterisation}D, when normalised to the static amplitude, the frequency response is fairly uniform. The increase in loading of the fin ray has minimal impact on the cutoff frequency, with a slight shift from 4.5 Hz for the unloaded case to 4 Hz under the heaviest weight. Resonance peaks emerge around 1-2 Hz, becoming more pronounced as the load increases. Increased weight of the loaded fin ray contributed to increased momentum, facilitating overshooting and resonances, as previously observed in Figure \ref{fig:singleactuatorchracterisation}C.


When embedded in a fin, the amplitude and frequency response are highly dependent on the material and geometry used. To exemplify the performance in a robotic system, a FOLD actuator was embedded in a square 33 x 33 x 0.8 mm silicone membrane. The added stiffness reduces the maximum amplitude achieved. The step response of the fin ray within the membrane, with a rise time of approximately 400 ms, confirms significant damping from water interaction (Fig. \ref{fig:singleactuatorchracterisation}C). The frequency response of the integrated fin shows a reduced cutoff frequency of 1 Hz due to water drag and damping effects (Fig. 1D).




When operated at 3 Hz, the presented actuator has a lifetime of more than a million cycles. It is noteworthy that at failure, no mechanical failure occurred in the PP, but the point of failure was the copper wire used for the coil. The region where the thin wires exit the coil is subjected to repeated stress due to coil movement, eventually leading to the breakage of the wire. This further demonstrates that PP is a suitable material choice for the fin ray as the deformation is well within the elastic region, and the forces are low enough to not introduce creeping.

\subsection*{Proof-of-concept integration of the fin ray: The CuttleBot}

A bioinspired aquatic robot was designed and fabricated to evaluate the performance of the fin rays in a robotic system. We explored a dual undulating fin design, previously investigated by \cite{zhou_design_2012, low_modelling_2009, pliant_energy_systems_robotics_2023,festo_bionicfinwave_2023, yin_kinetic_2021} and demonstrated impressive manoeuvrability. We leverage the small size of the FOLD actuator to explore combinations of frequency, wavelength and amplitude in a palm-sized robot.

\begin{figure}[htbp!]
    \centering
    \includegraphics[width=1\textwidth]{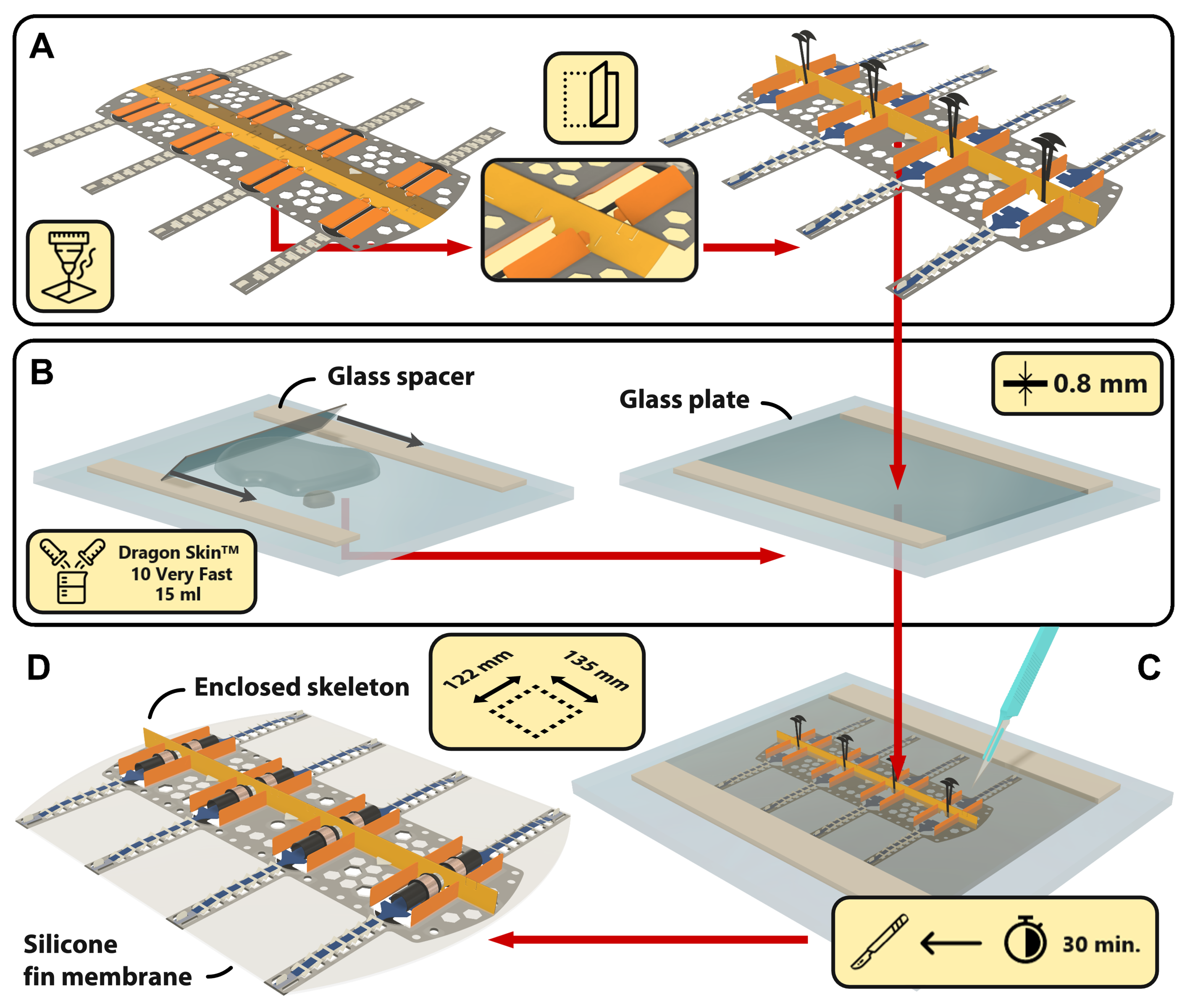}
    \caption{\textbf{Assembly of the CuttleBot.} \\{\color{gray}\small(\textbf{A}) Folding of the body: The laser-cut skeleton is folded to create spine and ribs, tendons are inserted through the T-hooks and secured to the fin rays. (\textbf{B}) Blade casting of the 0.8 mm thick silicon elastomer film. (\textbf{C}) The CuttleBot skeleton is embedded into the liquid silicone film, cured for 30 minutes and the outline defined using a scalpel. (\textbf{D}) The coils and magnets for the VCAs are added to complete the assembly.}}
    \label{fig:assembly}
\end{figure}

Inspired by cuttlefish, rays and skates, the CuttleBot has two medial fins that run along the side of the body. Undulating motion of the fins, generating a travelling wave, propels water enclosed within the wave and generates a net thrust forward \cite{low_modelling_2009,wang_experimental_2013,xia_design_2022,yin_kinetic_2021}. The design and assembly of the CuttleBot is depicted in Figure \ref{fig:assembly}. The CuttleBot measures 135 mm in length and 125 mm in width, consisting of eight single actuators in four pairs. Four fin rays extend from each side of the body and are connected by a flexible membrane to complete the fin. This monolithic construction allows for a sturdy structure and quick assembly. 

The assembly process consists of four main steps. Starting with a flat laser-cut skeleton structure consisting of PP that is folded to create the spine and ribs, similar to the singular fin ray (Fig. \ref{fig:assembly}A). After inserting and securing the tendons, the arrowhead limiters on each fin ray are bent upward (the folding steps are shown in Movie S3). In the next step, shown in Figure \ref{fig:assembly}B, two glass spacers are used for blade-casting/doctor-blading to create a 0.8 mm thick layer of silicone elastomer (Smooth-on Dragon Skin™ 10 VERY FAST) \cite{smooth-on_inc_dragon_2024} on a glass surface. The CuttleBot skeleton is then embedded into the silicone elastomer to cure (Fig. \ref{fig:assembly}C). In its liquid state, the silicone flows through cut-out holes in the skeleton and fin rays, enclosing them. After approximately 30 minutes of curing, the outline of the CuttleBot is cut with a scalpel and stencil before it is lifted from the glass surface. Excess silicone is removed from the cavities to make space for the VCAs. To assemble the VCA, the coils are attached to the tendons and the magnets to the spine, completing the CuttleBot (Fig. \ref{fig:assembly}D).

\subsection*{Characterisation of the CuttleBot}

\subsubsection*{Undulating fin characterisation}

\begin{figure}[htbp!]
    \centering
    \includegraphics[width=1\textwidth]{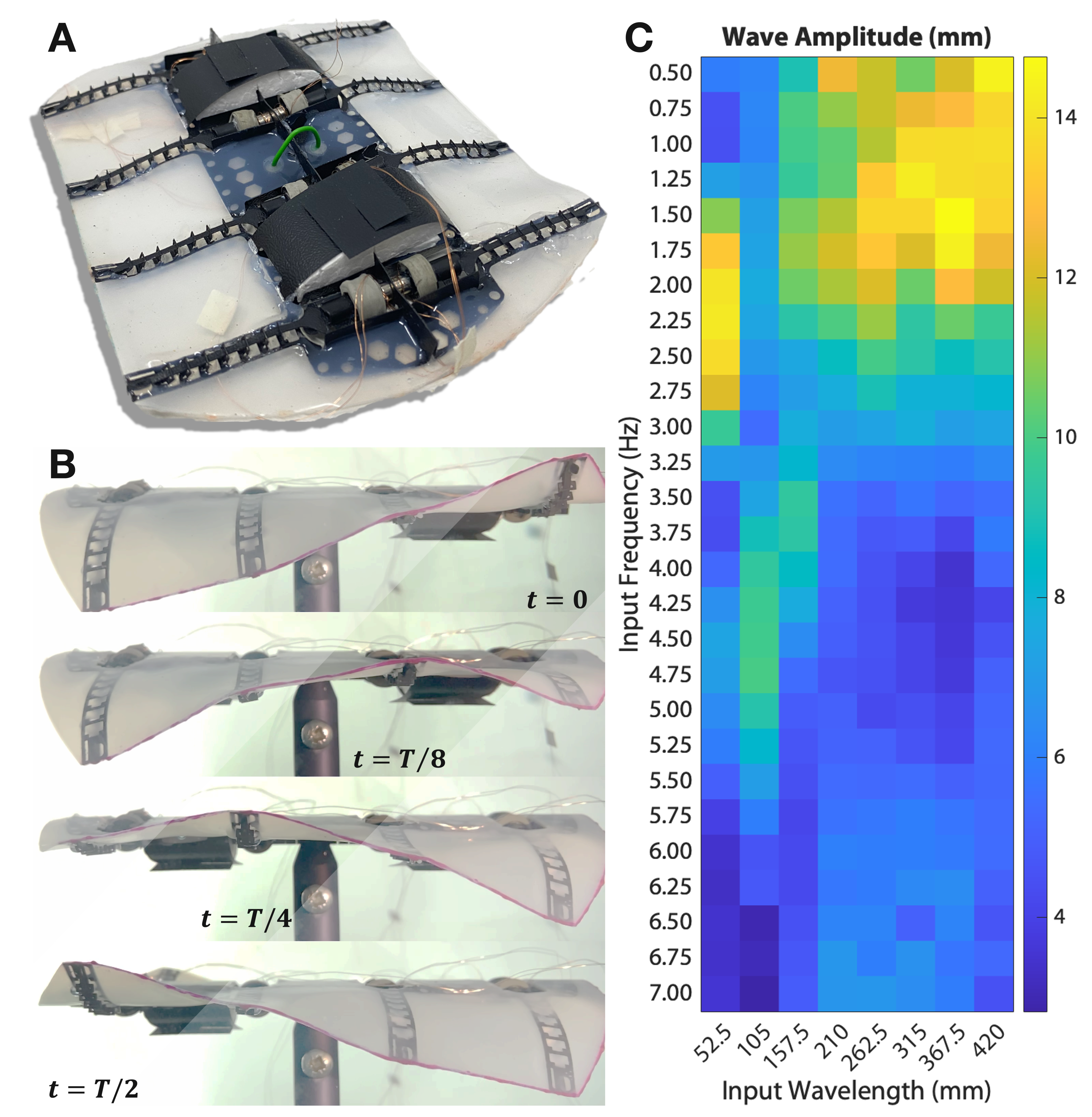}
    \caption{\textbf{Characterisation of the fin membrane.} \\{\color{gray}\small \textbf{(A)} Completed CuttleBot. \textbf{(B}) Side view of travelling wave on one fin. \textbf{(C)} Wave amplitude at wavelengths and frequencies combinations.}}
    \label{fig:fincharacterisation}
\end{figure}

Figure \ref{fig:fincharacterisation}A shows an image of a CuttleBot. Weighing 20~grams, it has been equipped with polystyrene floats, symmetrically positioned at the front and rear of the CuttleBot, to achieve close to neutral buoyancy. Similar to its biological counterpart, a travelling sine-wave is imposed on the fins on the left- and right-hand side of the CuttleBot, respectively, to facilitate swimming. The kinematics of an ideal travelling sine wave formed at the fin edges in terms of its amplitude $A(x,t)$ as a function of position $x$ along the fin edge and time $t$ can be described as

\begin{equation}
    A(x,t) = A_\text{max}(I) \ \sin(kx \pm \omega t)
\end{equation}

\noindent, where $A_\text{max}(I)$ is the maximum amplitude determined by the input current $I$ driving the VCAs, $k = 2\pi/\lambda$ is the wave vector as a function of the wavelength $\lambda$, $\omega$ is the angular frequency, and $\pm$ resembling a forward ($-$) and backward ($+$) travelling wave. Since the positions $x_i$ between the fin rays actuating the fin are determined by the design, the relative phase-offsets between the fin rays can be determined in a straightforward manner and each fin ray driven accordingly.

Figure \ref{fig:fincharacterisation}B exemplary shows half an actuation cycle of the fins of a stationary (clamped) CuttleBot in water. A travelling wave propagates from the right- to the left-hand side, indicated by the highlighted region. In order to investigate the kinematics in more detail, thresholding techniques were used to reconstruct the amplitude of the magenta dyed fin edge from recorded videos. The excitation wavelength is varied over a range of 52.5 to 420 mm, equivalent to 0.5 to 4 body (fin) lengths (BLs) and the frequency is varied from 0.5 to 7 Hz with increments of 0.25 Hz. The heat map in Figure \ref{fig:fincharacterisation}C shows that a maximum wave amplitude is achieved at a driving wavelength of $\lambda=$ 367.5~mm and a frequency of $f=$ 1.75~Hz. Frequencies greater than 2~Hz result in a drop-off of the amplitude due to hydrodynamic drag, in agreement with the observations for the singular fin ray (Fig. \ref{fig:singleactuatorchracterisation}). Similarly, wavelengths shorter than 157.5~mm also result in a drop-off of the maximum achievable amplitude. As the wavelength approaches the body length of the CuttleBot, the minimum sustainable wavelength is limited by the number of fin rays in the system. The Nyquist criterion dictates that a travelling wave of wavelength $\lambda$ requires base points not to be spaced more than ${\lambda}/{2}$ apart to resemble it correctly. Additionally, cross-coupling between the individual fin rays is increased for shorter wavelengths. Shorter wavelengths result in an increased phase offset between fin rays, i.e. a greater amplitude difference. This increased amplitude difference leads to greater stretching of the silicone membrane connecting adjacent fin rays, resembling the coupling of neighbouring fin rays via a spring that exerts a force between them proportional to the strain applied. Overall, both the number of fin rays and cross-coupling between fin rays limit the minimum wavelength that can be sustained and the maximum amplitude of the wave. This is further emphasised by the local maximum in amplitude formed at a wavelength of 52.5~mm and frequency of 2.25~Hz. Examination of the video footage of this peak reveals that a standing, instead of a travelling, wave is formed in the fin. This is attributed to the extreme difference in amplitude of adjacent fin rays and thus maximum cross-coupling, resulting in counter-propagating waves originating from both ends of the fin that interfere and form a standing wave.

\subsubsection*{Swimming characterisation}
The locomotion capabilities of the CuttleBot were systematically investigated. The ability to control each fin ray individually allows for different swimming modes and exploitations of various degrees of freedoms (DOFs). We investigated the effect of the fin kinematic on propulsion by inducing travelling sine waves along the robot's fins, varying in wavelength and frequency. 

\begin{figure}[htbp!]
    \centering
    \includegraphics[width=1\textwidth]{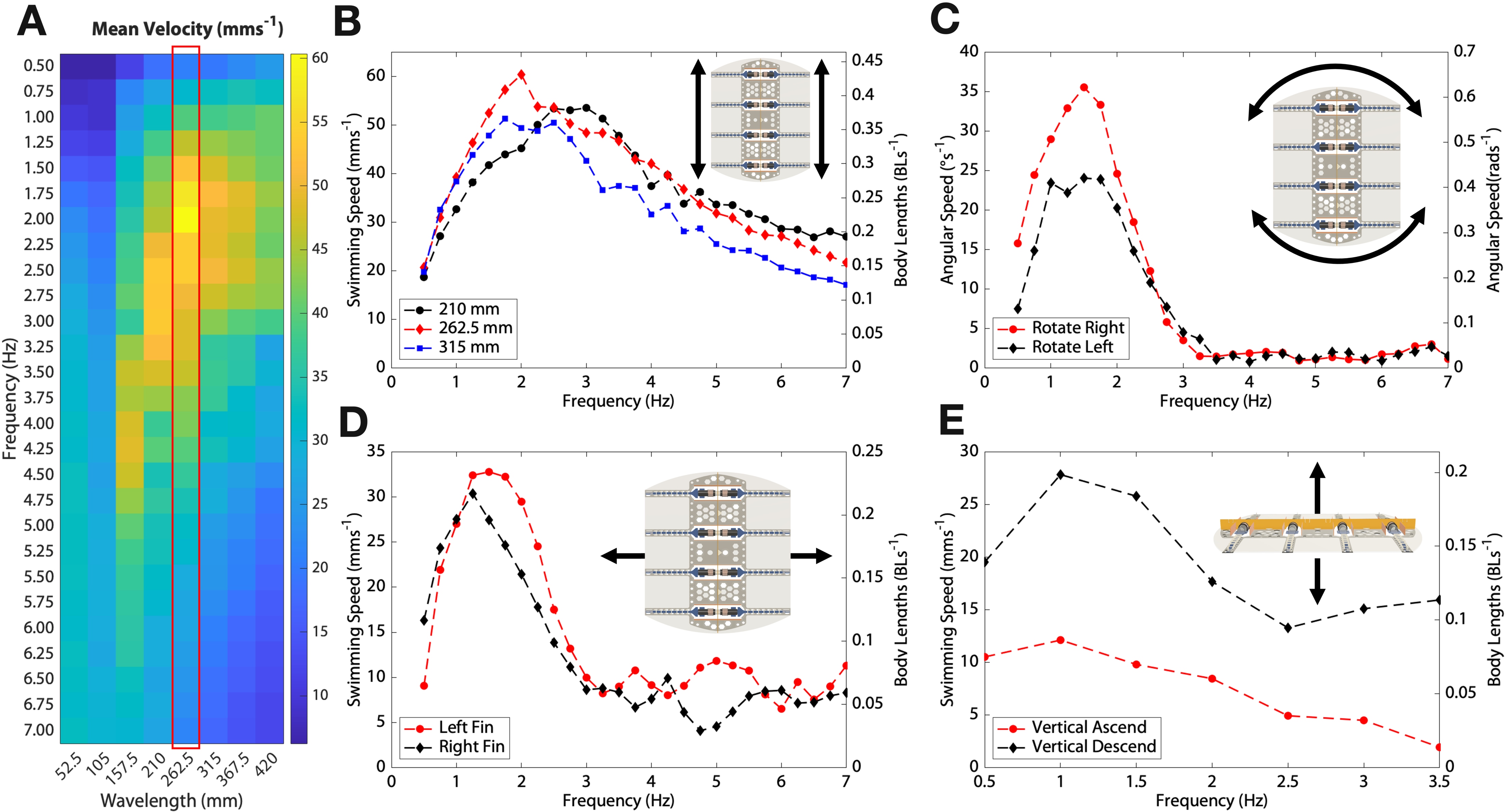}
    \caption{\textbf{Quantitative Swimming Characterisation.} \\{\color{gray}\small\textbf{(A)}: The mean horizontal swimming velocity of the CuttleBot at different frequency and wavelengths combinations; \textbf{(B)}: Peak horizontal(surge) swimming speed; \textbf{(C)} Peak rotation(yaw) speed; \textbf{(D)} Peak horizontal translation(sway) speed; \textbf{(E)} Peak ascending/descending(heave) speed.}}
    \label{fig:swim_char}
\end{figure}

When the sine waves on the fins were synchronized in phase, the CuttleBot moved in the body-fixed surge direction. The backward travelling waves led to forward movement and vice versa.  An automated video analysis system was used to track the position of the robot over time, from which we deduced the velocity corresponding to each combination of wavelengths and frequencies. Movie S4 shows the robot swimming at four different wavelength-frequencies combinations. As depicted in Figure \ref{fig:swim_char}A, the maximum mean velocity achieved was 60.4 mm/s (0.45 body lengths per second~(BL/s)) at a frequency of 2 Hz and a wavelength of 262.5 mm. Notably, the CuttleBot's velocity is adjustable over a range of 10 to 60 mm/s and the thrust profile is periodic (see Supplementary Materials), opening possibilities for precise operations such as controlled hovering in aquatic environments.

Diving deeper into the swimming dynamics, it was observed that velocity rose almost linearly with increased frequency up to a certain threshold where it reached a peak, beyond which it began to diminish (Fig. \ref{fig:swim_char}B), consistent with literature\cite{he_development_2015}. Notably, the peak fin amplitude (Fig. \ref{fig:fincharacterisation}C) does not coincide with the maximum swimming velocity (Fig \ref{fig:swim_char}A), suggesting a complex relationship between fin movement and thrust generation. The literature consistently states that thrust is expected to be directly related to the volume of water displaced by the fin within a given time frame\cite{smits_undulatory_2019,zhang_computational_2007,he_development_2015}. Theoretically, increasing the amplitude of the undulatory wave should maximise the enclosed volume of water displaced, just as elevating the frequency should accelerate the rate of displacement. However, a larger amplitude also increases drag due to a greater projected area, and the frequency is limited by the mechanical cutoff frequency of the fin\cite{blevins_rajiform_2012}. Therefore, achieving optimal propulsion requires a nuanced balance between these interrelated parameters.

A comparison with marine animals sheds further light on the observed horizontal swimming peak velocity of the bioinspired CuttleBot. Fish fins are often modelled as water foils and are described using a dimensionless figure called the Strouhal number $St = f\cdot A/U$, where $f$ is the frequency of oscillation, $A$ is the width of the wake (often approximated as the peak-to-peak oscillation amplitude of the fin), and $U$ is the swimming velocity\cite{eloy_optimal_2012,triantafyllou_optimal_1993,triantafyllou_hydrodynamics_2000,taylor_flying_2003}. This ratio could be thought of as the speed of the lateral movement of the fin versus the velocity produced from the thrust. Studies of biological species state Strouhal numbers 0.2~$\leq St \leq$~0.4 for oscillatory animals with swimming motion wavelength longer than body lengths\cite{triantafyllou_optimal_1993, sfakiotakis_review_1999, blevins_rajiform_2012}, and 0.6~$\leq St \leq$~2 for undulatory swimming at wavelength shorter than body length \cite{van_weerden_meta-analysis_2014}. This suggests an optimum frequency-amplitude combination. Another dimensionless figure is the specific wavelength $SW = \lambda / A_\text{mean}$, the ratio of the wavelength $\lambda$ of an undulation to its mean amplitude $A_\text{mean}$. The finding of \cite{bale_convergent_2015} is that there is a convergent evolution towards an Optimal Specific Wavelength (OSW) of 20, for speed optimisation (equivalent to maximising propulsive force), irrespective of frequency, length, height or overall shape of the fin. This has been observed for 22 species of fish of varying lengths (spanning two orders of magnitude), different body and fin morphology, and no geographical or taxonomical correlation and also substantiated experimentally with a robotic knife fish. The observed swimming behaviour of the CuttleBot is correlated with a Strouhal number of $St =$ 0.8 and a specific wavelength of $SW =$ 21.88 at peak velocity. These numbers are in good agreement with the Strouhal number range observed in marine animals and also the observed OSW. 


Beyond forward and backward swimming, the CuttleBot can yaw when the travelling waves on the left- and right-hand-side propagate in opposite directions. Angular momentum is induced, leading to rotation on the spot (zero-degree turning radius) (Fig. \ref{fig:swim_char}C). Keeping the wavelength fixed at $\lambda = 262.5$~mm, for which a peak in forward and backward swimming has been observed, variation of the frequency allows control of the angular speed with a peak of 35 $^\circ$/s observed at a frequency of $f = 1.5$~Hz. We attribute the difference in observed peak rotational speed for clockwise and counterclockwise rotations to the influence of the tethers powering the actuators.

Sway can be achieved by flapping either the left- or right-hand side fins. Sinusoidal operation of all fin rays on one side in phase leads to a side-ways motion of the robot, pushing the robot away from the fin that is flapping, shown in Figure \ref{fig:swim_char}D. Similar to horizontal forward/backward swimming and rotation, a peak in speed occurs at a frequency of $\approx$ 1.25~Hz. However, unlike previously described swimming modes, as the frequency increases, the motion of the actuator transitions from flapping to cupping (Fig. \ref{fig:singleactuatorchracterisation}A). In contrast to flapping, this cupping mechanism pulls the robot, producing thrust towards the same side as the fin. These two mechanisms compete, which results in two peaks, the first at 1.25 Hz indicates when the normal flapping motion is dominant, and the second smaller peak at 5 Hz when the flapping motion diminishes and the cupping motion takes over, pulling the robot.

Heave can be achieved by synchronously flapping the fins on both sides of the CuttleBot up- and down-wards (Fig. \ref{fig:swim_char}E). 
Ascending motion is realised by actively pulling the fins down from their horizontal plane and then allowing them to return passively to their neutral state when the actuators are switched off. The descent is facilitated by the inverse process. For these vertical movements, the fin rays are driven by a rectangular waveform with a 50\% duty cycle at varying frequencies, rather than the sinusoidal waveforms used for horizontal locomotion. A velocity peak is noted around 1 Hz for both ascending and descending, though the absolute values for each are not identical. This discrepancy can be attributed to the challenges in attaining perfect neutral buoyancy and inherent asymmetries in the up-and-downward fin strokes. These three additional swimming gaits can be seen in Movie S5.

\begin{figure}[htbp!]
    \centering
    \includegraphics[width=0.8\textwidth]{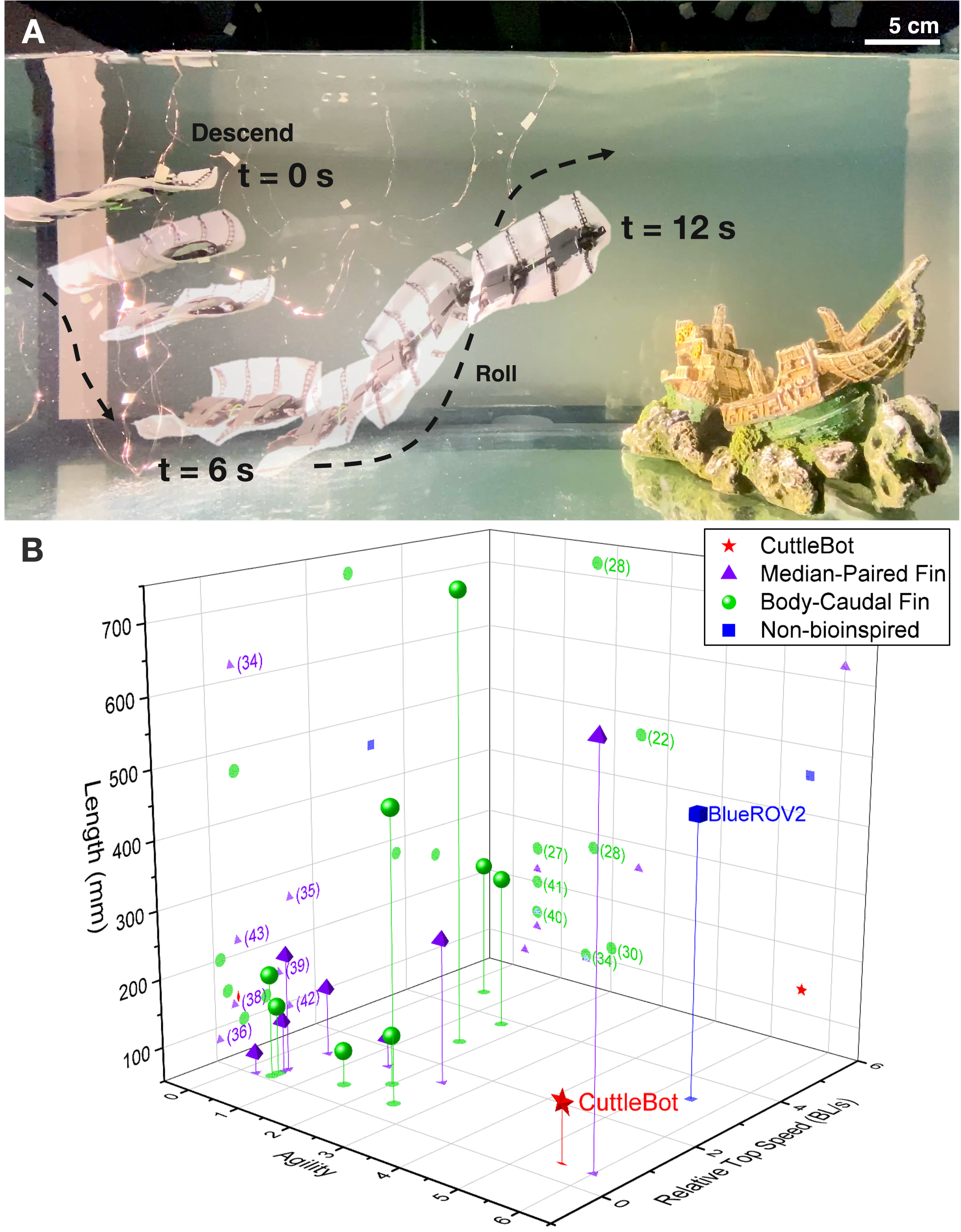}
    \caption{\textbf{Free Swimming and Comparative Performance} \\{\color{gray}\small \textbf{(A)} CuttleBot free swimming, the images are at two seconds intervals. First, it descends to the bottom of the tank, then swims diagonally upwards and rolls right. \textbf{(B)} The CuttleBot in relation to existing biomimetic and bioinspired aquatic robots.}}
    \label{fig:comparison}
\end{figure}

Beyond the four independent DOFs demonstrated, the CuttleBot can perform complex trajectories and dynamic manoeuvres by transitioning between these isolated swimming modes. For instance, as illustrated in Figure \ref{fig:comparison}A and Movie S6, the CuttleBot initially descends by synchronously flapping its fins, then shifts to forward motion near the tank's bottom through sinusoidal excitation. At T = 6 s, altering the phase shift from $\Delta \phi = 0^\circ$ to $\Delta \phi = 20^\circ$ between the left and right fins prompts the CuttleBot to roll. This manoeuvre illustrates the critical relationship between instability and manoeuvrability \cite{weihs_stability_2002}, specifically when generating rotation around the anteroposterior body axis as observed by \cite{sumikawa_changes_2022}.

Figure \ref{fig:comparison}B provides a comparative analysis of recent advances in bioinspired aquatic robots, focusing on their size, top speed, and agility. The proposed agility metric for aquatic bioinspired robots evaluates multiple dimensions of locomotive proficiency beyond degrees of freedom (the scoring methodology is detailed in the Supplementary Materials). The ubiquitous BlueROV2 from BlueRobotics offers a practical performance benchmark, capable of traversing five degrees of freedom independently and swims at 1.5 m/s (3.3 BL/s) \cite{blue_robotics_inc_bluerov2_2022}. The combination of performance, configurability, and affordability have rightfully justified its popularity in aquatic robotics research. In comparison, robots modelled after BCF locomotion (green) are noted for their remarkable top speeds; the TunaBotFlex \cite{white_tunabot_2021} swims at 4.7 BL/s. However, they exhibit limited agility, typically confined to forward movement with a relatively large turning radius due to the propulsion mechanism of BCF motion. In contrast, robots inspired by median-paired fin (MPF) locomotion (purple) theoretically offer greater agility, but experimental demonstrations have shown limited ranges of motion and degrees of freedom (DOFs). A notable exception is the design by \cite{yin_kinetic_2021}, which uses two preloaded rubber fins with a permanent sinusoidal shape, moved by a series of brushless motors. By varying the phase difference between the motors, the robot can independently traverse four DOFs and even move over land. However, this design imposes limitations on the fin's workspace (wavelength and amplitude) as well as the size, complexity, and cost of the robot. The CuttleBot contribute to this diverse landscape by demonstrating similar maneuvrability to \cite{yin_kinetic_2021} at one-quarter the size. Leveraging the small size of the FOLD actuator, four independent actuators can be fitted, which allows the fin to adopt various arbitrary shapes.

\subsubsection*{Power and efficiency characterisation}

The experiments presented were performed at a peak current of 212~mA per VCA to avoid amplitude clipping of the fin rays, as determined for a singular fin ray (Fig. \ref{fig:singleactuatorchracterisation}B). In this configuration, the entire CuttleBot consumes approximately 11.8 W of electrical power, which we refer to as 100\% power consumption in the following. Every fin ray of the CuttleBot had been operated under identical conditions, i.e. each VCA has been supplied with the same power. However, since the fin rays driving each fin constitute a coupled system, it is of interest to understand how an uneven power distribution influences the performance of the CuttleBot. 

\begin{figure}[htbp!]
    \centering
    \includegraphics[width=1\textwidth]{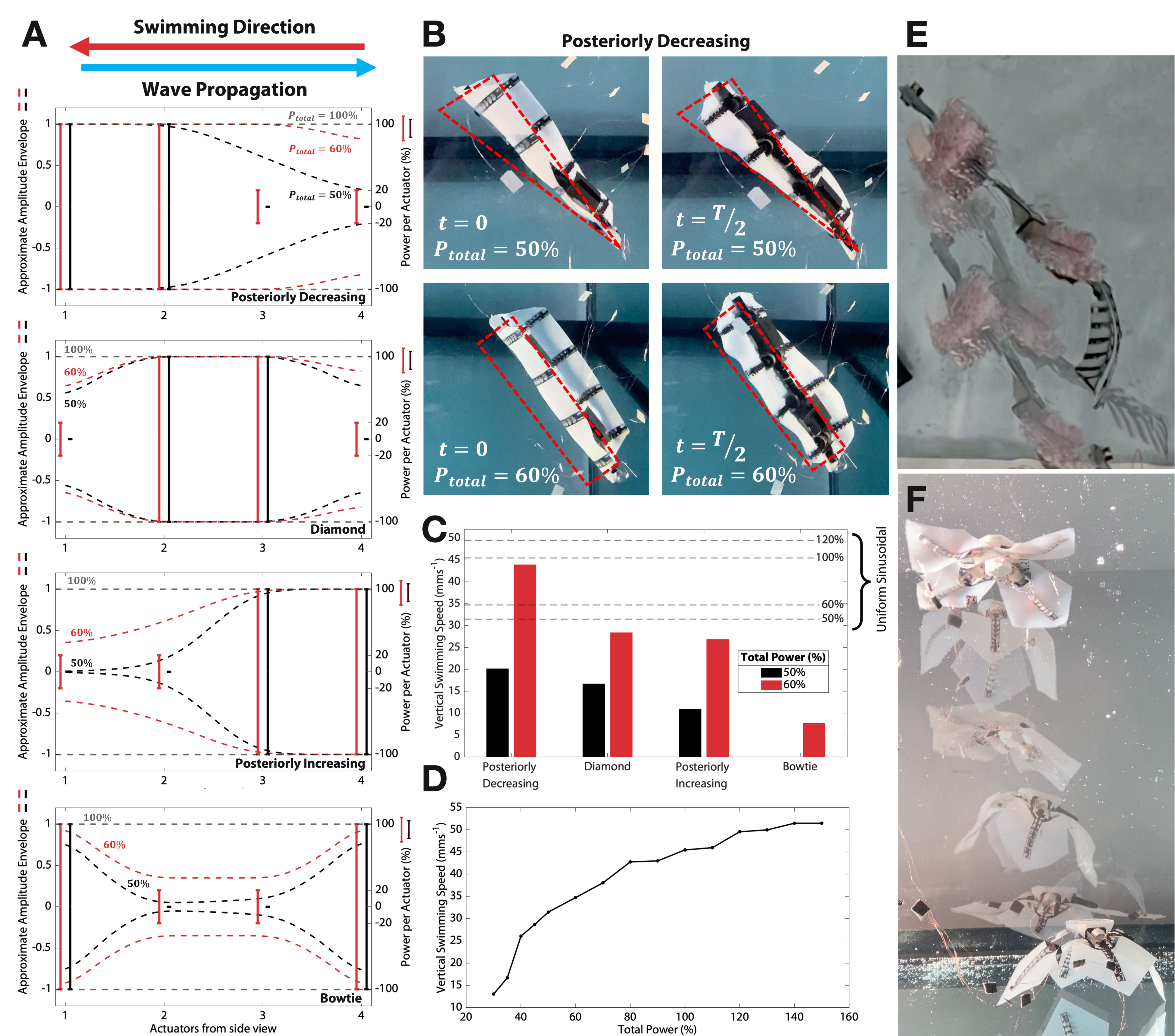}
    \caption{\textbf{Power Velocity Relationship.} \\{\color{gray}\small\textbf{(A)} Power distribution at each fin ray and the approximate resulting wave envelope. \textbf{(B)} Posteriorly decreasing wave envelope at 50\% and 60\% total power. The wave envelopes are outlined in red. \textbf{(C)} Swimming performance of each wave envelope benchmarked against standard uniform sinusoidal swimming. \textbf{(D)} Uniform sinusoidal swimming speed. \textbf{(E)} A tuna-inspired fish robot with undulating tail. \textbf{(F)} An ephyra-stage jellyfish robot with four flapper.}
    \label{fig:power}}
\end{figure}

By disabling (0\% power per fin ray) or partially powering (20\% power per fin ray) different pairs of fin rays, the fin can form various wave envelopes (Fig. \ref{fig:power}A and Movie S7). We evaluate and compare their effect on vertical swimming performance. In the first experiment, the two pairs of fin rays at the back of the CuttleBot were disabled. The amplitude at the anterior was unaffected but we observed a noticeable reduction in the posterior amplitude. This formed a wave envelope that is posteriorly decreasing. However, partially powering the two fin rays at the back leads to an almost complete recovery of the envelope, indistinguishable to that of uniformly powered fin rays at 100\% power consumption. Figure \ref{fig:power}B shows frames of the CuttleBot operating at an overall power of 50\% and 60\%, respectively, corresponding to the cases of both fin rays at the back completely disabled and partially powered, highlighting the difference in wave envelopes during locomotion. In the other cases, where the first and last fin rays are underpowered (diamond-shaped), the first two (posteriorly increasing), and in the centre of the fin (bowtie-shaped), this recovery is less pronounced.

The difference in the wave envelopes formed translates into the swimming speed. Figure \ref{fig:power}C shows the swimming speed for the four different cases of posteriorly decreasing envelope, diamond-shaped envelope, posteriorly increasing envelope, and bowtie-shaped envelope at 50\% and 60\% total power, respectively. The swimming speed for a uniform sinusoidal envelope at different total power levels is shown as a benchmark for comparison. Surprisingly, the posteriorly decreasing envelope at 60\% total power results in swimming speeds almost comparable to that of a uniformly powered sinusoidal wave. The others perform significantly worse than their uniform sinusoidal envelope counterparts. The bowtie-shaped wave performs the worst, unable to lift the robot from the floor.


At this stage, the mechanism behind the observed phenomenon is not fully understood and is beyond the scope of this study. However, we speculate that for our case of a posteriorly decreasing envelope, the wave induced by the fin rays at the front propagates towards the fin rays at the back, facilitated by the surrounding water and formed vortices, requiring little excessive energy for amplification, akin to keeping a pendulum or children's swing in motion. 

Limited research in this area has not arrived at a general consensus on the effect of wave envelope on thrust and efficiency \cite{he_development_2015,zhang_computational_2007,hu_hydrodynamic_2022}. The CuttleBot constitutes a promising platform for studying the correlation of localised power distribution and waveforms/envelope with the overall performance of the biomimetic system, which is a challenging task in biological species.


\subsection*{Versatility in morphology}

Lastly, we wish to emphasise the versatility and integration capability of the presented actuator through the integration into two other bio-inspired robotic systems; the T.U.N.A. (Tendon-driven UNdulating tAil) and the jellyfish. In Figure \ref{fig:power}E, the T.U.N.A robot has a pronounced caudal fin powered by two actuators, which oscillate side-to-side to provide forward thrust achieving a maximum speed of 48.5 mm/s (0.34 BL/s) at 2.75 Hz. The jellyfish has four fins that can be powered individually to achieve control over more than one DOF (Fig. \ref{fig:power}F). It is capable of vertical swimming at 15.6 mm/s (0.14 BL/s). As such, the demonstrated actuator clearly exhibits potential for integration into future biomimetic and bioinspired aquatic robotic systems.

\section*{CONCLUSION}

In this work, we have demonstrated a bioinspired hybrid actuator with active curvature control that emulates the functionality of a fish fin ray, enabling complex manipulation of a fin. FOLD actuators can be rapidly fabricated by laser cutting ordinary polypropylene film and hand-assembled in minutes, resulting in an accessible and rapid manufacturing process. The cost of materials is low (approximately £0.80/\$1), and the actuators can be controlled with simple electronics. Additionally, the design allows FOLD actuators to be integrated into various robotic systems, as demonstrated with the CuttleBot, T.U.N.A., and the jellyfish robot. We highlighted the CuttleBot, where multiple actuators were embedded and collectively actuated to create 3D deformations of two fins, smoothly transitioning between propagating wave patterns, gliding, and flapping. Current approaches mostly utilize the flexibility of the fin ray structure to achieve some degree of passive curvature \cite{bianchi_design_2022,liu_fin_2017}, or neglect this aspect completely by modeling the fin ray as a stiff rod \cite{siahmansouri_design_2011,low_modelling_2009}. We improve upon this by actively controlling the curvature of the fin. Moreover, the CuttleBot qualitatively and quantitatively demonstrated and supported biological observations such as the Strouhal number\cite{taylor_flying_2003,eloy_optimal_2012} and Optimal Specific Wavelength\cite{bale_convergent_2015,nangia_optimal_2017}. The FOLD actuator and the presented platform hold high potential for the study of the principles of biomimicry and robotics, serving as an ideal tool for instructional purposes, enabling students and researchers to directly engage with these concepts.

Future work will focus on optimising the shape of the actuator to achieve the desired deflection characteristics, ultimately working towards automated inverse design. Numerous opportunities exist to explore new materials and configurations to enhance the efficiency and control of the actuator, as well as to enable new modes of actuation. Additionally, we aim to investigate the CuttleBot’s application in evaluating and improving energy efficiency within biosinpired systems, and to develop its system modeling through a combination of fluid-structure interaction and computational fluid dynamics. The simplicity of the CuttleBot’s deployment and the ease of data collection offer great potential for fine-tuning simulations governed by differentiable physics \cite{ma_diffaqua_2021,du_underwater_2021,lee_aquarium_2023}. Efforts will also be directed towards refining the manufacturing process to establish a catalogue of readily deployable robotic models for on-site use. The next phase of development anticipates the CuttleBot’s transition to wireless autonomy, setting the stage for fully autonomous operation.

\section*{MATERIALS AND METHODS}
\subsection*{Materials and Constructions}

Sheets of polypropylene (PP) (Value 400135897 A4 Display Folder - Black, Cartridge People, UK) of thickness 0.2 mm have been laser cut with a laser cutter (Neje Master 7W [2~W optical power]). The black colour of the PP facilitates the absorption of the laser light with a wavelength of 450 nm. The PP sheets are glued with a thin layer of liquid, water-soluble PVA glue onto transparent acrylic (Perspex) sheets of 3 mm thickness. The high transparency of the acrylic sheets in the visible wavelength range prevents interaction with the laser cutter. Adhesion of the PP sheets to the acrylic carrier prevents buckling and curling of the PP sheets. Further, cut-outs in the PP sheets remain in place during the laser-cutting process. Engraving is used to create folds in the PP sheet with a laser power of 9$\%$ and 9~ms dwell time. A laser power of 30$\%$ and a dwell time of 30ms are used to fully cut through the PP sheet. After the cutting process, parts can be lifted from the acrylic plate using a scalpel or a pair of tweezers. The parts were then cleaned with water in an ultrasonic cleaner and a brush to remove glue and any remaining debris.

Black drinking straws with a diameter of 5~mm (Polypropylene Bendy Drinking Straws, Buzz Catering Supplies, UK) were mounted on a 3D-printed stage to fully suspend them into the air for laser cutting. During the cutting process, the top and bottom surfaces of the straw will be cut. The suspension of straws leads to thermal decoupling from the underlying surfaces and promotes enhanced symmetrical cutting of the top and bottom surfaces. A laser power of 9$\%$ ($\approx$ 0.18 W) and a dwell time of 9 ms is used for engraving. A laser power of 30$\%$ ($\approx$ 0.6 W) and a dwell time of 30~ms with two passes are used to cut the straws.

Washers laser cut from PP sheets were secured to the straw with small amounts of superglue (Loctite SuperGlue Liquid Precision Max) to assemble a bobbin for the coil, Fig. \ref{fig:singleactuator}C. The bobbin is slid onto a rod that is mounted in a low RPM hand drill. Subsequently, an enamelled copper wire of 100 $\mu$m diameter (BNTECHGO 38 AWG Magnet Wire, Amazon) is wound on the bobbin using the hand drill and distributed manually along the length of the bobbin. Approximately 11 m of copper wire is used for a single bobbin, where 9 m is wounded and 1m is slack at each end. The slack wires are secured together along the length with tape to prevent entangling. Standard 2.54 mm pitch 2-by-1 header pins are soldered onto the ends of the wires for electrical connection. The enamel of the wires evaporates during the soldering process at a temperature of 350\textdegree{}C, i.e. it is not necessary to remove the enamel beforehand.


After assembling the body, a thin layer of Dragonskin\textsuperscript{TM} 10 Very Fast is casted onto a glass plate employing a blade-casting/doctor-blading approach. Glass spacers with a thickness of 0.7~mm are secured to the plate using 0.1~mm double-sided tape on the left- and right-hand side allow producing a silicone film of 0.8~mm thickness. The body of the fish is then immersed into the silicone layer using tweezers. Cut-outs defined in the skeleton allow the silicone to fully enclose the PP sheets, ensuring a stable mechanical connection between the silicone and PP skeleton. After curing the silicone at room temperature, the outline of the fish is cut with a scalpel, using the stencil mask outline defined during the laser cut. The CuttleBot is then removed from the glass plate.


Overall assembly times totalled 3.5 hours, with 2.5 hours spent on making the coils, 20 minutes folding the body and attaching the tendons, and 40 minutes to mix and form the silicone around the body.


\subsection*{Electronics and Firmware}
The actuator is driven by an L298N H-bridge, controlled by pulse width modulation (PWM) signals from an Arduino Mega 2560. The sine values are stored in a lookup table and are modified to become the PWM values.

\subsection*{Experimental Setup}
Experiments were conducted within a glass aquarium of dimensions 100 cm by 40 by 50~cm. All parameters, barring voltage and current, were measured optically. For timing and synchronisation, we used a setup with three LEDs (see Supplementary Materials), which emitted flashes at designated moments within the actuation sequence or upon alteration of any variable (such as amplitude, frequency, wavelength, etc.).

The video recordings were made with an iPhone 11 Pro Max with the \textit{Yamera} app, which enabled customisation of shutter speed and ISO settings to minimise motion blur. Videos of single actuators were shot at 240 frames per second, ensuring at least 16 frames per actuation cycle for the maximum tested frequency of 15 Hz. This high frame rate and the stark contrast between the dark actuators and the white backdrop facilitated straightforward segmentation through intensity thresholding in MATLAB. For comprehensive fin characterisation, one edge of the fin was tinted magenta to stand out distinctly, enabling detection during processing via binary masking of specific colour values. When characterising free swimming, the Physlets Tracker software \cite{brown_innovative_2009} was utilised to track a group of moving pixels---manual placement of tracking points are sometimes required due to occlusions caused by water reflections, and/or distortions from surface waves and motion blur. Detailed methodology is provided in the Supplementary Materials.

\clearpage

\bibliography{references.bib}

\noindent\textbf{Acknowledgments:} The authors would like to thank Que Anh Dang, Rawan Elsayed, Kinjiro Amaro, and Andrew West for providing valuable feedback on the early drafts of this paper. The icons used in the figures were created by Smashicons, GregorCresna, NT Sookruay, Freepik, surang, JunGSa from www.flaticon.com. \\
\textbf{Funding:} Minh Vu acknowledges support by the UK Engineering and Physical Sciences Research Council (EPSRC) [Grant number EP/T517823/1]. \\
\textbf{Author Contribution}: 
M.V. and T.J.E. conceptualised the actuator and cuttlefish robot, and developed the fabrication method.  M.V., T.J.E, S.W. and A.W. developed the design, electronics, control software and visualised the data.  M.V. performed the experiments and collected the data. T.J.E., M.V., and C.M. conceptualised the jellyfish robot. R.R. further improved the design and developed the control, electronics, and performed experiments. A.M. and M.V. conceptualised, designed, and experimented with the fish robot. M.V. and T.J.E. wrote the initial draft of the manuscript. All authors contributed to the manuscript, provided feedback, and approved the final draft. T.J.E., S.W. and A.W. were responsible for the overall research supervision and objectives. T.J.E. secured the funding.\\
\textbf{Competing Interest:} All author declares no competing interest.\\
\textbf{Data and materials availability:} Design files and source code can be found at \\
https://github.com/nhatminh2h/FOLDer



\section*{Supplementary material}

# Supplementary Materials for
# Fin ray-inspired, Origami, Small Scale Actuator for Fin Manipulation in Aquatic Biomimetic Robots

**Minh Vu**[1,2*]**, Revathy Ravuri**[1]**, Angus Muir**[1]**, Charles Mackie**[1]**, Andrew Weightman**[3,4]**, Simon Watson**[1,4]**, Tim J. Echtermeyer**[1,2,5*]

[1]**Department of Electrical and Electronic Engineering, The University of Manchester**
[2]**Photon Science Institute, The University of Manchester**
[3]**School of Engineering, Faculty of Science and Engineering, The University of Manchester**
[4]**Centre for Robotics and Artificial Intelligence, The University of Manchester**
[5]**National Graphene Institute, The University of Manchester**
**Manchester, M13 9PL, United Kingdom**
*]**Correspondence: minh.vu@manchester.ac.uk,**
**tim.echtermeyer@manchester.ac.uk.**

# 1 ACTUATOR AND CUTTLEBOT MANUFACTURING & CONTROL HARDWARE

## 1.1 Preparation of plastic sheets

The 200 $\mu$m thick, black polypropylene (PP) sheets are glued to a transparent, 3 mm thick acrylic (Perspex) plate using liquid PVA glue. For this, a brush or tissue is employed to thinly and evenly spread the PVA glue over the PP sheet. Then, the PP sheet is laid on the acrylic sheet, pressed on, and any trapped air bubbles removed by squeezing/rubbing them towards the edges of the PP sheet to ensure intimate contact between the PP sheet and the acrylic.

## 1.2 Laser cutting

**We wish to remind users to adhere to relevant Health & Safety procedures and risk assessments for laser cutting. High intensity laser light can cause eye damage. Further, fumes/vapours are generated when laser cutting plastic materials, requiring sufficient**

**ventilation.**

As an example, the laser cut pattern for the backbone, tendon, and coil for a pair of actuators is shown overlaid in Figure 1A, and the design for the CuttleBot shown in Figure 1B. The backbone and body are first engraved following the yellow pattern, then cut using the purple pattern. On a separate PP sheet, the tendons are cut using the cyan pattern.

For designs that have both an engraving and cutting layer, we split the designs into two separate dxf files, one for the engraving and one for the cutting layer, respectively. The files can then be uploaded separately to the laser cutter for processing after each other. Alignment of the different layers with respect to each other is facilitated by small alignment marks (rectangles) in the outer corners of the layers. This enables the use of the out-of-the-box software with limited functionalities provided with the laser cutter (Neje Master 7W), particularly for less experienced users of laser cutters. Alternatively, the designs are straightforward compatible with more advanced software tools for the control of the laser cutter.The employed laser cutter operates at a wavelength of 450 nm. As such, we intentionally choose materials of black colour to facilitate laser light absorption by the materials.

The fabrication time for the laser cutting is in the order of 10 minutes for a pair of actuators and $\approx 40$ minutes for a full CuttleBot on our system.

The full designs for both the actuator pair and the full CuttleBot can be found on GitHub.

### 1.3  Folding Assembly

Using a scapel, the laser cut skeleton and tendons are lifted from the acrylic plate (Fig. 2A). Residual PVA glue is washed off using warm water and can be complemented with ultrasound cleaning to remove debris. Similarly, the acrylic plates can be cleaned and recycled for subsequent use.

The ribs are then folded upwards (Fig. 2B), before the spine is folded (Fig. 2C). Subsequently, the spine and ribs can be secured and joined by adhesive-less joining principles employing "click" principles (Fig. 2D). In the next step (Fig. 2E), the hooks of the fin rays are bent upwards, allowing the tendons to be inserted and joined with the tips of the fin-rays (Fig. 2F). To minimise residual stress in the body, each fold is performed back and forth (positive and negative 90°).

### 1.4  Silicone casting

After the assembly of the PP skeleton, a 0.8 mm thick layer of silicone is prepared on a glass surface, employing a blade casting approach with spacers to define the desired thickness (Fig. 3A). We elected to use Smooth-on Dragonskin$^{\text{TM}}$ Very fast 10 due to its flexibility/stretch-ability

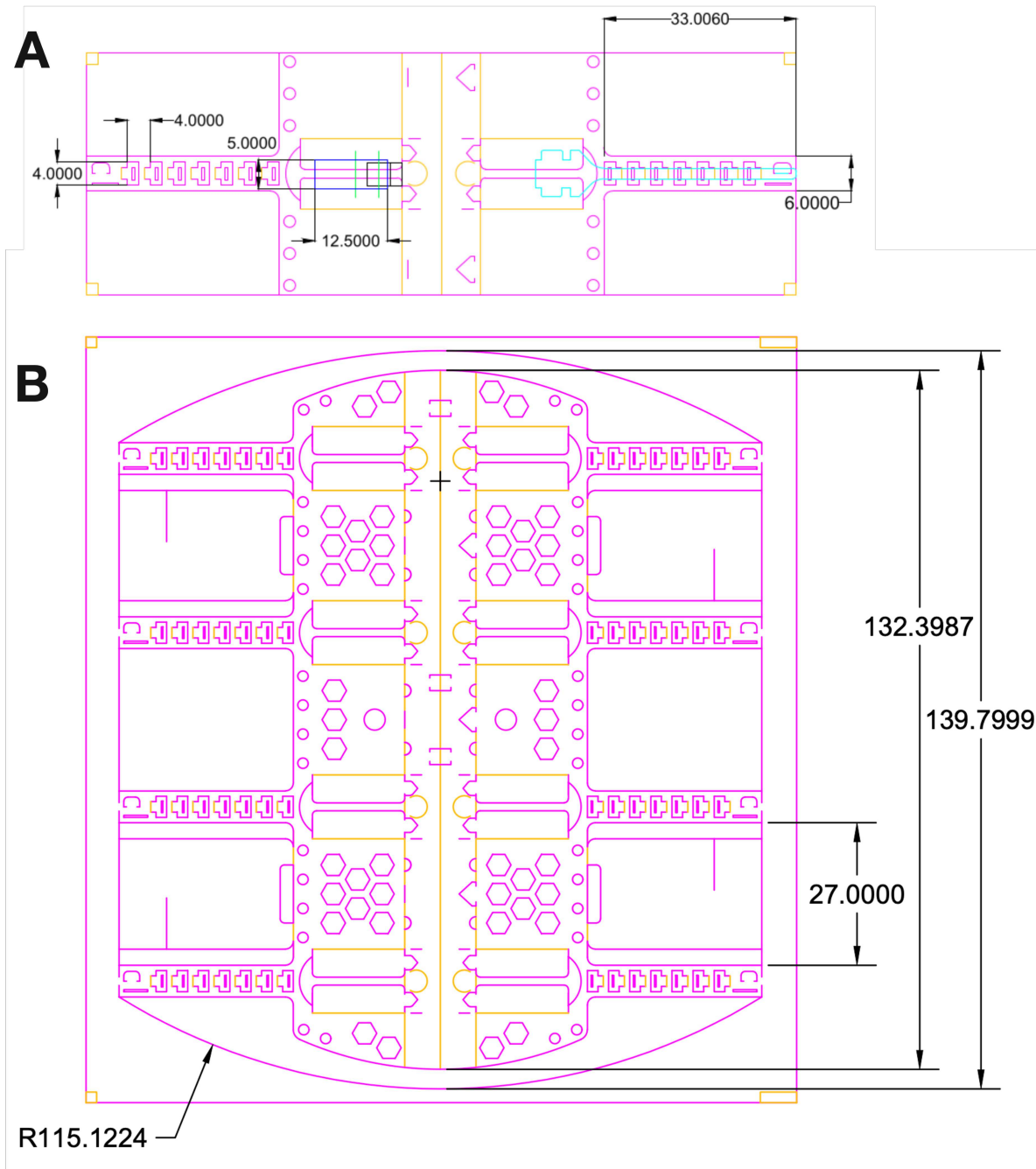

**Figure 1. Overlaid layers of laser cut drawing**

*(Purple) Cut lines for the body. (Yellow) engrave lines. (Cyan) Tendon. (Dark Blue) Straw Outline. (Green) Straw engrave lines. (Black) 3 Magnets.*

**(A)** Laser cut drawing for a single FOLD actuator. The tendon and straws are overlaid and aligned. **(B)** Laser cut drawing for a CuttleBot.

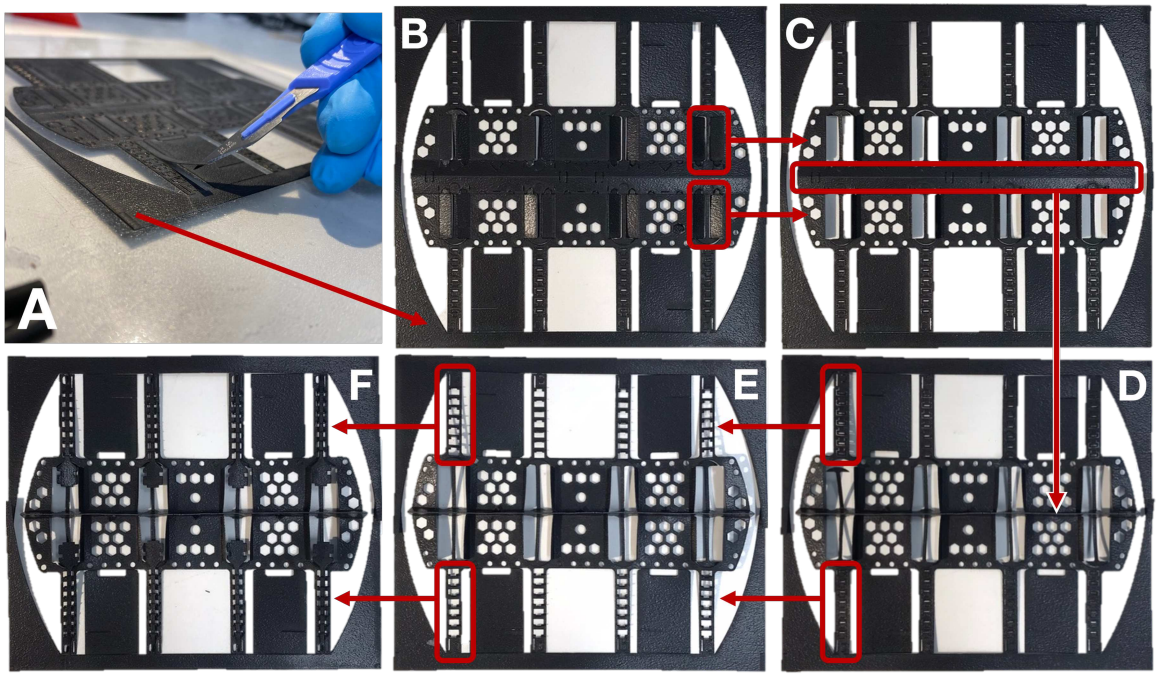

**Figure 2. The CuttleBot skeleton is folded from flat laser-cut polypropylene sheet.**
The red boxes highlight the areas of focus. **(A)** The polypropylene skeleton is lifted from the acrylic using a scalpel. **(B)** Top down view of a fully released skeleton. **(C)** The ribs are folded up. **(D)** The spine is folded up and connected to the ribs. **(E)** The fin ray loops are folded up. **(F)** Tendons are inserted through the loops.

(Shore Hardness 10A), high elongation before breakage (1000 %), density close to that of water (1.07 g/cm³), and fast cure time (30 minutes) (59). Subsequently, the skeleton is immersed into the silicone while it is still in its liquid form (Fig. 3B). The process is facilitated by gently "poking" the skeleton, e.g. with a wooden tooth pick, from above to ensure the silicone flows through all the holes and openings in the skeleton, fully enclosing the skeleton. Due to the fast cure time of the silicone, the skeleton will settle in the silicone film within a few minutes. Simultaneously, care needs to be taken to work cleanly, i.e. it should be avoided to cover the hooks of the fin rays and the tendons with silicone.

After the silicone is cured, the outline of the CuttleBot can be defined with a scalpel, removing any residual silicone outside the area defined by the stencil mask integrated into the laser-cut design (Fig. 3C).

Figure 3D shows an image of a finalised CuttleBot after the silicone has cured and the CuttleBot was carefully lifted/peeled off the glass substrate.

## 1.5 Fabricating voice coil actuators

In the first step, black plastic straws of 5 mm diameter are laser-engraved and -cut (see design on GitHub). Washers, laser cut from 200 $\mu$m thick PP sheets are secured to the laser-cut straw with small amounts of superglue (Loctite SuperGlue Liquid Precision Max) to assemble a bobbin for the coil (Fig. 4A). The bobbin is slid onto a rod that is mounted in a low RPM hand drill (Fig. 4B). Subsequently, an enamelled copper wire of 100 $\mu$m diameter (BNTECHGO 38 AWG Magnet Wire, Amazon) is wound on the bobbin using the hand drill and manually distributed along the length of the bobbin. Approximately 11 m of copper wire is used for a single bobbin, where 9 m is wound and 1 m is slack at each end (Fig. 4C). The wire is secured to the bobbin using tape, e.g. copper tape (Fig. 4D). The slack wires are secured together at equidistant points along their length (approximately every 15 cm) with a small tape ribbon to prevent entangling. Standard 2.54 mm pitch 2-by-1 header pins are soldered onto the ends of the wires for electrical connection. The enamel of the wires evaporates during the soldering process at a temperature of 350°C, i.e. it is not necessary to remove the enamel beforehand. In the last step, the tendon can be slid into the bobbin, with a self-aligned latching mechanism securing it (Fig. 4E)

## 1.6 Bill of Materials

Commonly, the cost of materials depends on quantities purchased. We have selected quantities for the bill of materials (BOM) that are reasonable for the purchase of smaller labs, i.e. not too small quantities that would drive up the cost per unit but also not unreasonably large quantities.

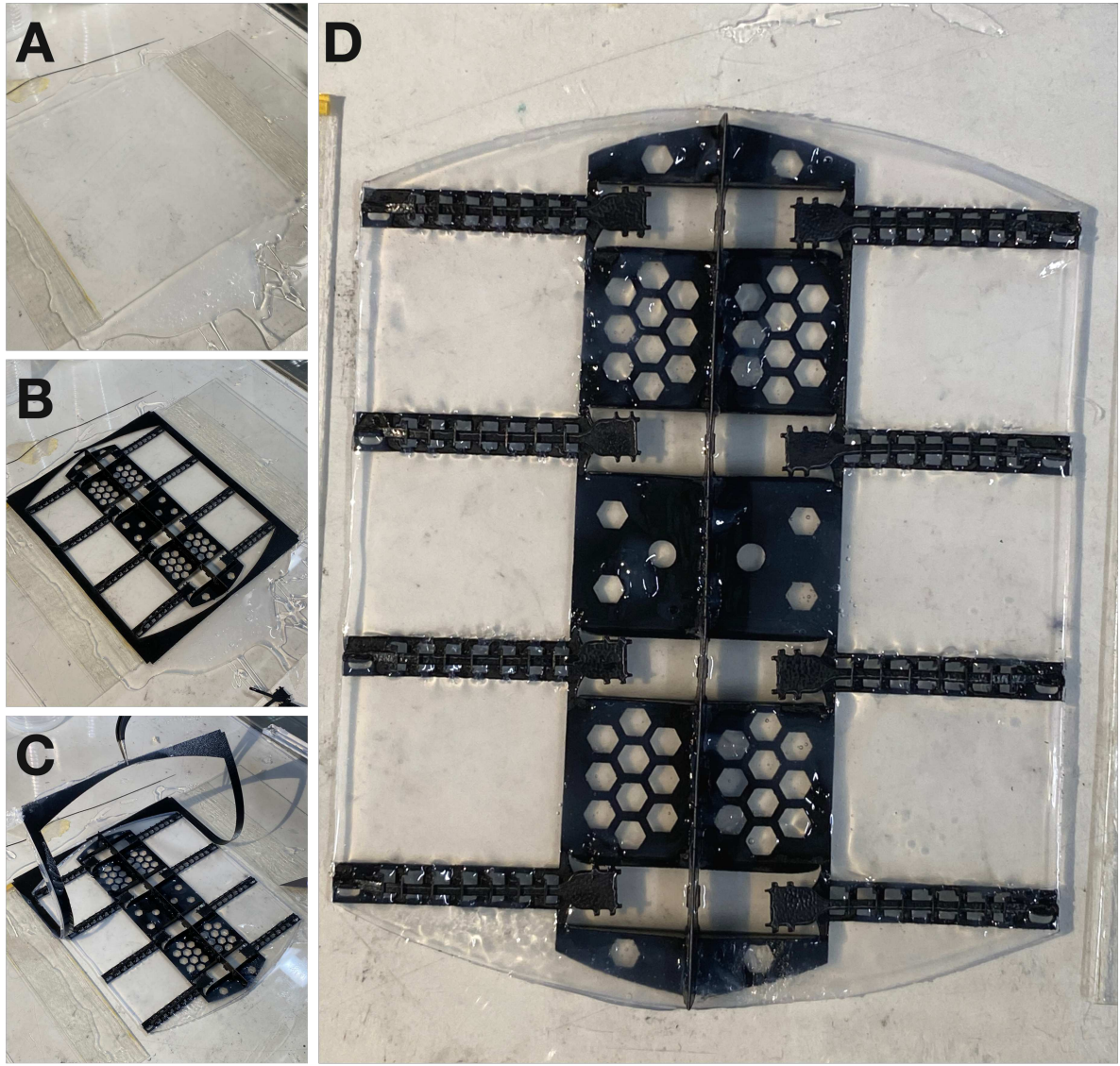

**Figure 3. Enclosure of the CuttleBot skeleton in silicone.**
**(A)** 0.8mm thick Dragonskin 10 is blade-casted onto a glass surface. **(B)** An assembled skeleton is immersed into the liquid silicone. **(C)** After 30 minutes, the silicone cures and excess parts are removed with a scalpel. **(D)** Finalised CuttleBot, ready for assembly and integration of VCAs and magnets.

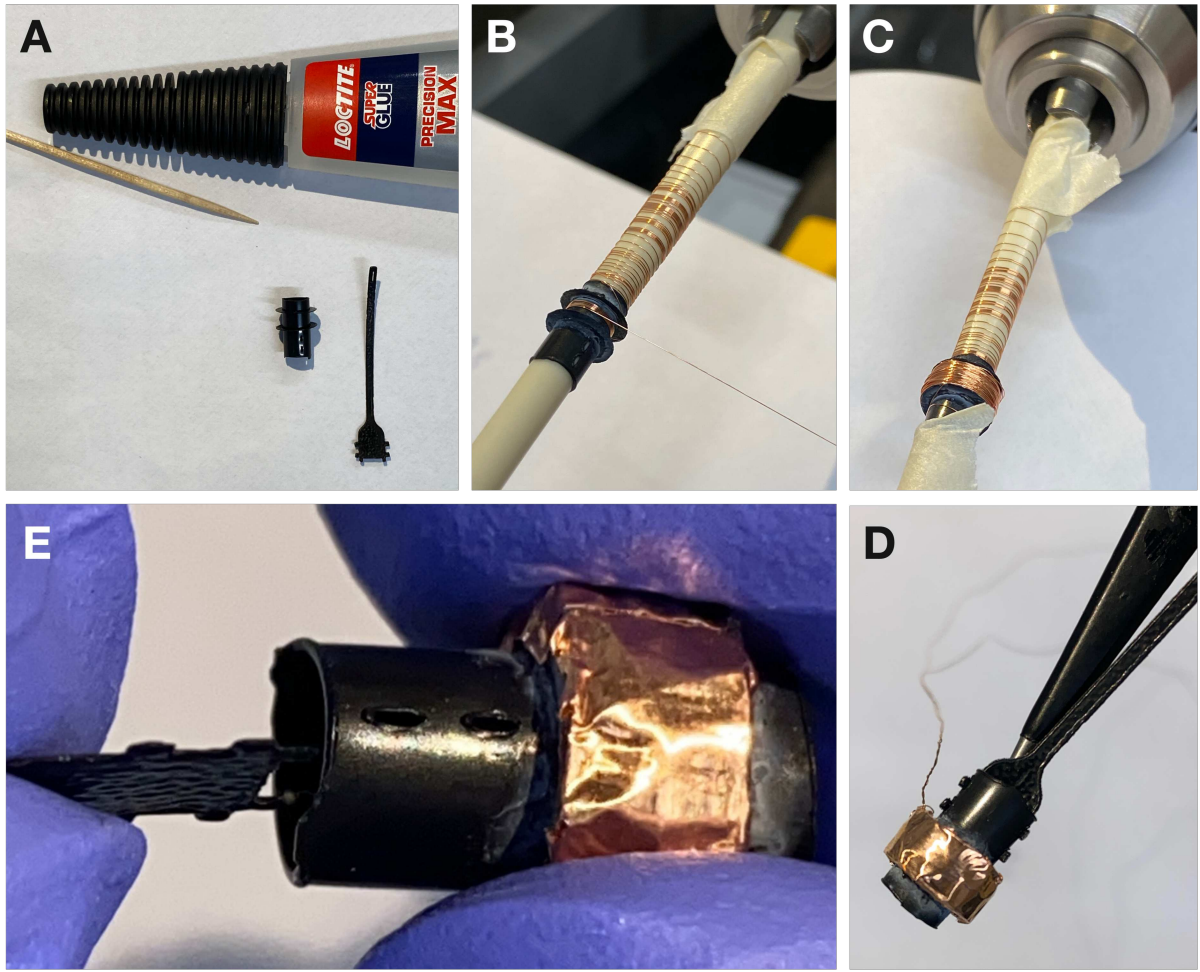

**Figure 4. Fabrication of VCAs**
(A) Loctite SuperGlue is used to attach the rings to the laser cut straw (B) The coil bobbin is mounted onto a drill and copper wires are wounded (C) A fully wounded coil (D) Tape is put over the coil to secure the winding. (E) The tendon is attached to the VCA.

Based on these, the BOM shows that the material costs for a single FOLD actuator are £0.80 ($1.00) and that of a full CuttleBot £6.50 ($8.12) (Table 1).

# Table 1. Bill of Materials

| Item | Minimum Purchase Amount | Unit | Cost of Item | Cost per Unit | Single Actuator Quantity Used | Single Actuator Cost (£) | CuttleBot Quantity Used | CuttleBot Cost (£) | Link |
|---|---|---|---|---|---|---|---|---|---|
| Polypropylene Sheet (A4) | 25 | sheet | £4.19 ($5.24) | £0.168 ($0.210) | 1 | £0.17 ($0.21) | 1 | £0.17 ($0.21) | https://www.cartridgepeople.com/Stationery/Value-400135897-A4-Display-Folder-Black-25-Pack-21166.html |
| Neodymium N52 Magnet (4mm diameter) | 100 | magnet | £15.00 ($18.75) | £0.150 ($0.188) | 3 | £0.45 ($0.56) | 24 | £3.60 ($4.50) | https://www.guysmagnets.com/neodymium-magnets-c11/guys-magnets-4-mm-x-2-mm-n52-high-grade-neodymium-disk-p408 |
| Polypropylene Straw (5mm diameter) | 250 | straw | £1.55 ($1.94) | £0.006 ($0.008) | 1 | £0.01 ($0.01) | 8 | £0.05 ($0.06) | https://www.buzzcateringsupplies.com/polypropylene-bendy-drinking-straws-8-inch-20cm-black-pack-of-250.html |
| Enamel coated copper wire 0.1mm diameter (4 oz) | 1615 | m | £18.30 ($22.88) | £0.011 ($0.014) | 11 | £0.12 ($0.16) | 88 | £1.00 ($1.25) | https://www.amazon.co.uk/BNTECHGO-AWG-Magnet-Wire-Transformers/dp/B0823C4R2S/?th=1 |
| Loctite SuperGlue (10 g) | 10 | g | £4.80 ($6.00) | £0.480 ($0.600) | 0.1 | £0.05 ($0.06) | 0.8 | £0.38 ($0.48) | https://www.diy.com/departments/loctite-precision-max-liquid-superglue-10g/5010305061229_BQ.prd |
| Smooth-on DragonSkin 10 VeryFast (2 pint/1140 ml) | 1140 | ml | £37.00 ($46.25) | £0.032 ($0.041) | 0 | £0.00 ($0.00) | 40 | £1.30 ($1.62) | https://www.benam.co.uk/dragon-skin-10-very-fast |
| Total Cost | | | £92.46 ($115.58) | | | £0.80 ($1.00) | | £6.50 ($8.12) | |

### 1.7 Control Electronics

Figure 5 shows a block diagram of the control hardware. An Arduino Mega was chosen as micro-controller due to the large number of available output pins. The Arduino controls motor drive boards (L298N), one per two actuators, via pulse width modulation (PWM) to control the magnitude and polarity of the current supplied to each VCA. This allows independent control of each VCA/fin ray of the CuttleBot. The total cost for the control electronics is less than £50 ($65).

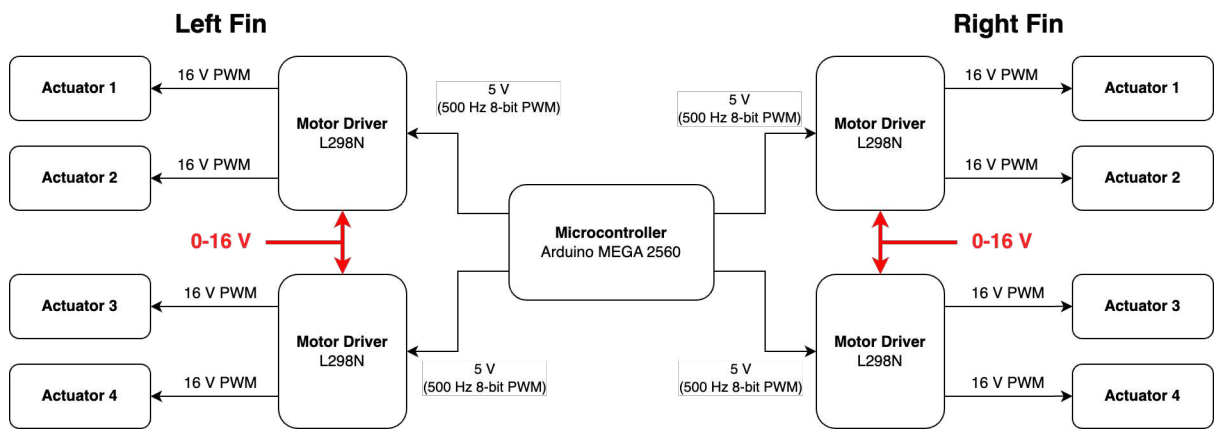

**Figure 5. Block diagram of the control electronics for the CuttleBot.**

## 2 SINGLE ACTUATOR CHARACTERISATION

The following section describes in detail the experimental setup and subsequent data analysis of the characterisation of a single fin ray-inspired origami electromagnetic tendon-driven (FLOP) actuator.

### 2.1 Voice Coil Actuator Force Measurement

The operation of a Voice Coil Actuator (VCA) is fundamentally governed by the Lorentz force, which describes the force exerted on a current-carrying conductor in a magnetic field. When current flows through the VCA's coil, it interacts with the magnetic field from a permanent magnet, resulting in a force that drives the actuator. For theoretical modelling, we employed the model from ref (*51*), which considers the geometry of the coil and the magnetic field characteristics.

For experimental characterisation of the force created by the VCA, six magnets are attached

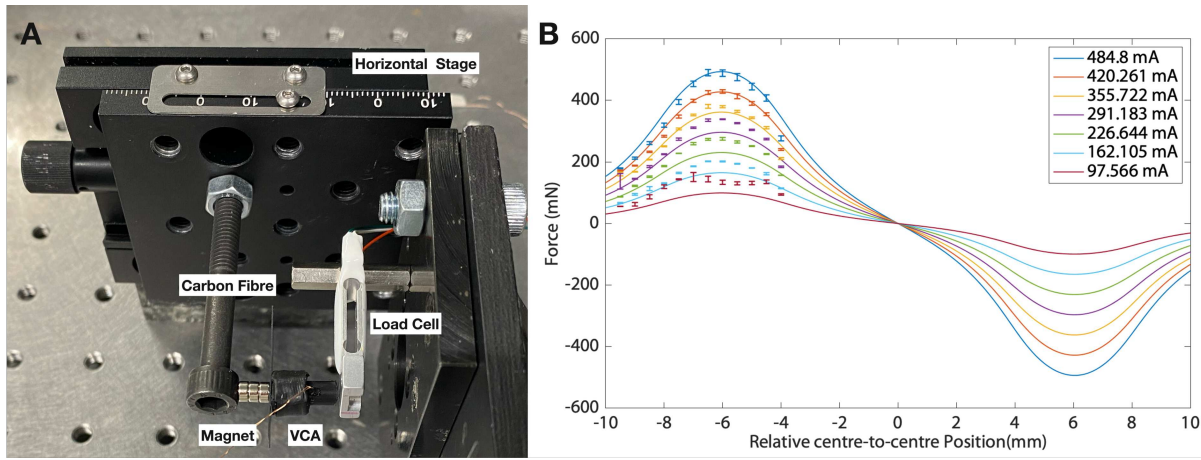

**Figure 6. VCA Force Measurement**

**(A)** Experiment Setup. **(B)** Measurements of the force as a function of relative alignment between coil and magnets and current, compared to simulation results.

to the flat side of a square M6 nut which is fixed to a horizontal stage with a screw (Fig. 6A). The magnets are bifurcated with a thin piece of carbon fibre to emulate the setup of a single actuator in which three magnets are mounted to each side of the spine. A load cell (Phidgets 3139_0) is fixed to the opposing end. The coil is slid on the magnets, sandwiched between the carbon fibre and the strain gauge. When the coil is electrically excited, it pushes on the strain gauge. Voltages of (16, 13.3, 10.6, 7.9, 5.2) V are applied to the coil and the force is recorded; the data from the load cell is read-out using a Wheatstone bridge (Phidgets Wheatstone bridge) with a sampling frequency of 113 Hz to mitigate noise from the mains supply, the PWM actuation signal, and other electronics. As additional parameter, a linear stage allows varying the distance between the magnets and strain gauge, i.e. the relative alignment between the coil and the magnets. In this experiment, we varied the alignment over a range of 5.5 mm with a step size of 500 $\mu$m. The resulting experimental data is the VCA's force as a function of driving current and relative alignment of coil and magnets, which is in good agreement with above theoretical model (Fig. 6B).

## 2.2 Optical Measurement of Deflection

Video recordings were employed to evaluate the mechanical deflection/changes of actuators under excitation. All videos were recorded on an iPhone 11 Pro Max using the Yamera app to control the shutter speed. The fin ray was black and the background was cream white, which when gray-scaled provides high contrast for image segmentation. We found that a simple adaptive

threshold mask provided the best extraction, compared to edge detection techniques (e.g. Canny, Sobel, LoG...). After the binary mask is applied, smaller blobs and noise are filtered out setting a lower threshold for their area, leaving the fin ray isolated.

To facilitate automated analysis of video recordings of experiments with large parameter space, e.g. variations in excitation magnitude or frequencies, a set of LEDs has been recorded alongside the fin ray. The LEDs are synchronised with the control code for the actuator, and are electrically controlled by the Arduino. In this way, the LEDs can be used to indicate changes in parameters controlling the actuator. An example are more complex experiments characterising a full fin of the CuttleBot under sinusoidal excitation: if both the wavelength and frequency need to be varied, the top, red LED could indicate when the frequency changes, and the middle, green LED, when the wavelength changes; both LEDs emitting a short flashing signal at each change, respectively.

The MATLAB script starts by prompting the user to define the regions where the LEDs are, the region where the fin ray deflects, and the root of the fin ray (Fig. 7A). In the area where the fin ray deflects, a mean adaptive threshold with a sensitivity setting of 0.2 is applied, resulting in a binary image. Small noise and unwanted artefacts are then removed using morphological opening (eroding then dilating). Any remaining blobs are filtered out by minimum area and perimeter threshold, which cleans up the image (Fig. 7B). Subsequently, the boundary of the tip of the fin ray is traced in red (Fig. 7C), and the tip of the fin ray (marked in green) is identified as the pixel farthest from the root (marked in red) (Fig. 7D).

## 2.3 Weighted fin rays and fin rays with hydrodynamic load

To allow loading of the fin ray with additional weights at the tip, a modified version of the fin ray with a hole at the tip has been designed (Fig. 8A). The hole can be employed to attach M3 screws and nuts (Fig. 8(B)), to which additional weights can be added (Fig. 8C). An M3 plastic screw is used and has been cut to a length of 3 mm.

To evaluate the effect of hydrodynamic loads on the fin ray, a version of a singular fin ray with a silicone fin has been designed (Fig. 8D). A rectangular frame (feature width 1.5 mm) around the fin ray enables structural integrity of the silicone membrane, i.e. it prevents collapsing/flapping of the silicone fin during actuation while still allowing following the curvature dictated by the fin ray.

All designs can be found on GitHub.

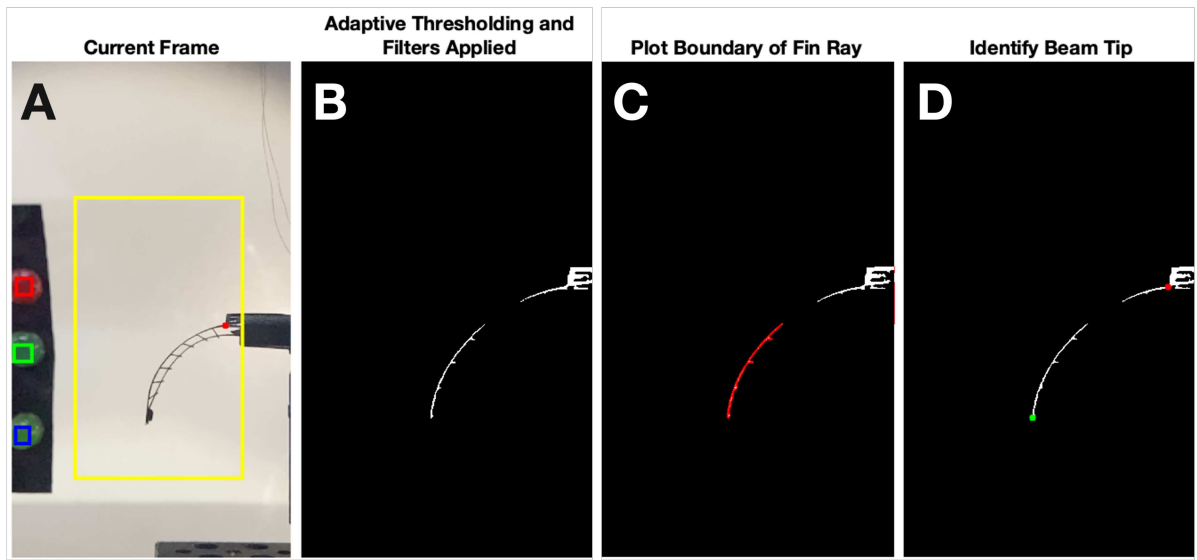

**Figure 7. Optical measurement of deflection**

**(A)** The current frame marked with predefined regions of interest: three LEDs on the left (red, green, blue rectangles), the area where the fin ray operates (yellow rectangle), and the root of the fin ray (red marker). **(B)** Adaptive thresholding and filtering applied, resulting in a black and white image of the fin ray. **(C)** Boundary points are drawn around the fin ray. **(D)** The tip (green) and root of the fin ray (red) are determined.

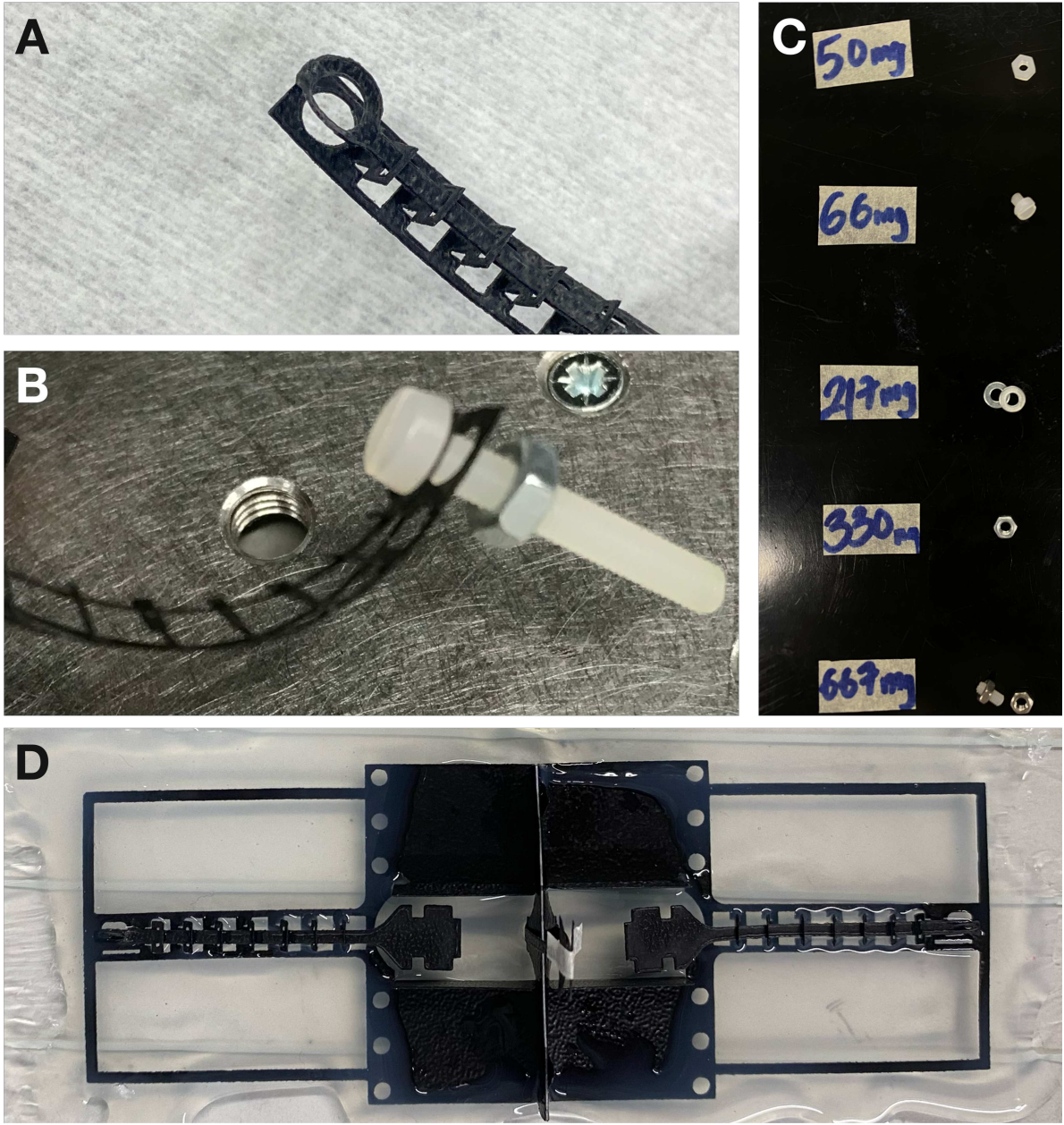

**Figure 8. Weights attached to the tip of the fin ray**
(A) Modified fin ray with 3 mm diameter hole at the tip. (B) Exemplary depiction of fin ray with screw and nut. (C) Employed weights (from top to bottom): Plastic nut, plastic screw, two metal washers, metal nut, and two metal nuts and plastic thread. (D) Singular fin ray with silicone fin.

# 3 CUTTLEBOT FREE SWIMMING CHARACTERISATION

The following section describes the experimental setup and data analysis of the swimming characterisation of the CuttleBot.

## 3.1 Control of fin kinematics

A flowchart of the control strategy is depicted in (Fig. 9). The waveform of the desired travelling wave to be generated on both the left- and right-hand side fins of the CuttleBot

$$A(x,t) = A_{\max}(I) \ \sin(kx \pm \omega t) \tag{1}$$

depends on numerous input parameters, depicted in green: the frequency, wavelength, phase offset $\theta$ between the fins, travel direction of the wave (forward/backward), and the envelope of the wave. To generate the desired sine wave for the CuttleBot's movement, a lookup table (LUT) containing 512 predefined amplitude values is employed. The phase difference between the two fins $\theta$ and the desired wavelength is used to calculate the phase difference between the fin rays $\phi$. Based on the phase difference $\phi$, the index for which value needs to be retrieved from the LUT can be determined for each fin ray. Further, the desired direction of wave propagation dictates the order in which the values are accessed from the LUT, i.e. counting up or down. Each fin ray's amplitude is then scaled using a multiplier to shape the wave's envelope. The frequency of the wave dictates the rate at which these amplitude values are updated, effectively controlling the speed of wave propagation through the CuttleBot's fins.

By modifying a combination of these wave parameters, various degree of freedom can be achieved. Defining the "head" of the CuttleBot at fin ray 1 (FR1), forward surge is achieved by sequentially accessing the LUT, generating a sinusoidal wave propagating from FR1 to FR4 synchronously across both fins. Backward motion is achieved by reversing the direction in which the LUT is accessed, the sinusoidal waves propagate from FR4 to FR1. To yaw in both directions, the direction of one fin is reversed; the robot turns towards the side that is reversed. To translate from side to side (sway), all the fin rays on one fin are operated at 0° phase difference, while the other side is turned off. The fin rays are actuated synchronously with a sinusoidal signal, equivalent to flapping one fin. To rise vertically, the fins flap from neutral to the bottom most position, and vice versa for sinking, this limit is achieved by multiplying the input signal by half and shifting the wave below or above the neutral line. Rolling can be achieved by operating the two fins on the left- and right-hand side at a 30° phase difference,

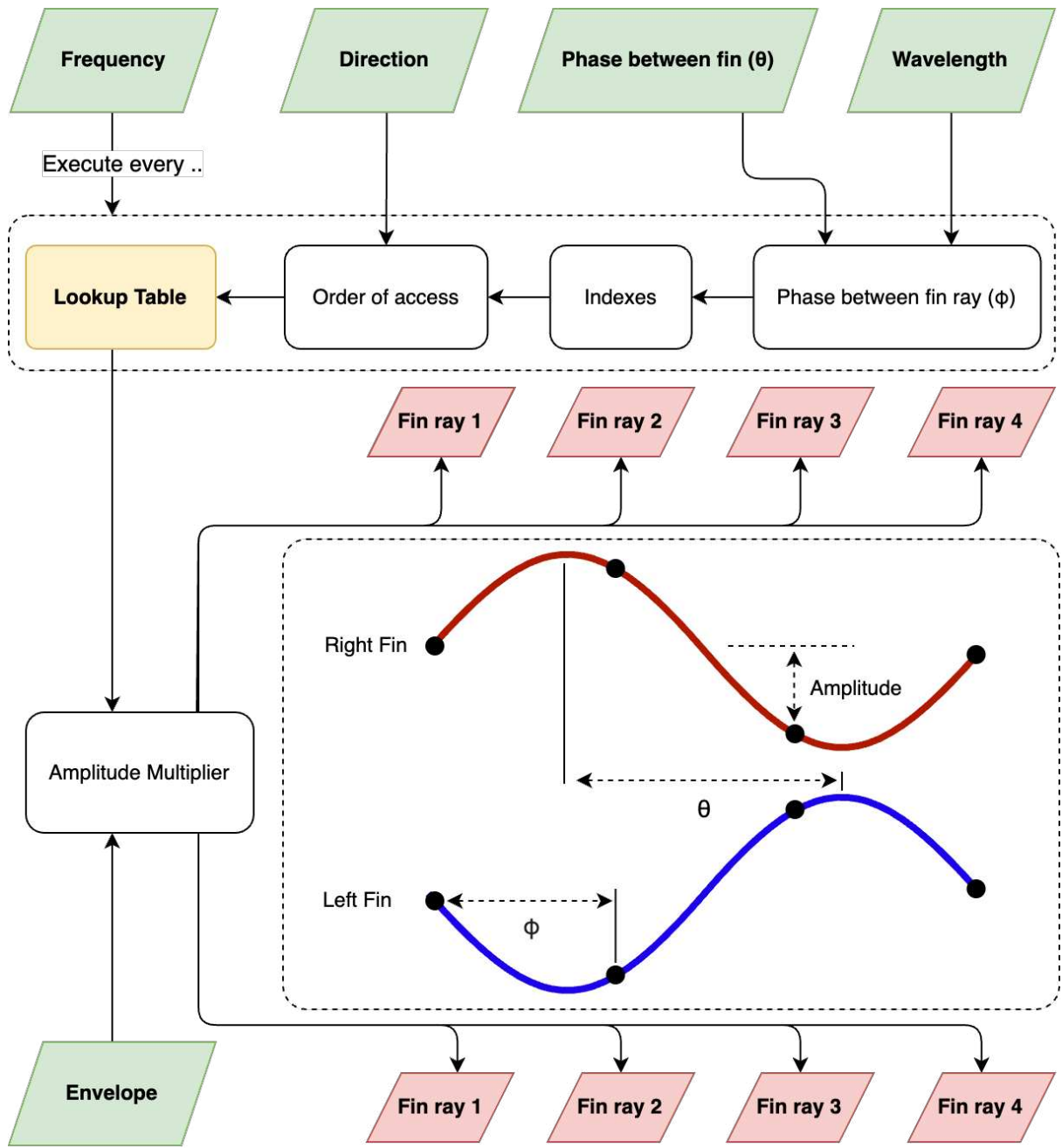

**Figure 9. Driving signal generation**

The driving signals for the fins (4 fin rays on each side) is generated from the desired frequency, envelope, direction, wavelength, and phase offset between fins on the left- and right-hand side of the CuttleBot (green). The inset shows example waves with a wavelength of one body length for the left- and right-hand side, respectively. The black circular markers represent the tip positions of the fin rays. In this example, the two fins are 180° out of phase.

## 3.2 Tracking swimming performance

The CuttleBot is set up to swim in a fish tank with dimensions 1000 mm long, 400 mm wide and 500 mm deep. A green wire is attached to the centre of the CuttleBot, acting as a marker, which can be tracked using Physlet Tracker (*72*).

Generally, the CuttleBot is programmed to swim in a predetermined reciprocal motion, i.e. forward then backwards, turn right then turn left, etc with varying combinations of wavelengths and frequencies. For example, for the straight swimming experiment, the CuttleBot is programmed to swim forward and backward at a fixed wavelength, and after each cycle the frequency is incremented. An example for data extracted with Physlets tracker is shown in Figure 10. Here, straight swimming has been performed with a sine wave of wavelength of 150 mm and increasing frequencies from 0.5 to 14 Hz, and the extracted displacement $y$ and velocity $v_y$ are shown in the top and bottom panel in Figure 10A, respectively. The displacement increases and decreases periodically, reflecting the forward- and backward-motion of the CuttleBot. The periodicity decreases over time due to the increasing driving frequency. In this way, the velocity can be extracted for each wavelength-frequency combination. If required, noise and artefacts in the data stemming from ripples on the water surface disturbing the video recording, have been smoothed using a Savitzky-Golay filter with a polynomial degree of three. Similarly, the characteristics of other locomotion patterns can be analysed, for example rotation (Fig. 10B).

## 3.3 Comparative Analysis of Biomimetic Aquatic Robots

A baseline score is established by awarding points for the robot's controlled degrees of freedom (DOFs), recognising both the presence and the range of each DOF. Half a point is allocated for each DOF, with an additional half point added for bidirectional control within the same DOF. A quarter-point penalty is imposed for robots that operate at fixed speeds, acknowledging the limitations imposed on agility by the inability to modulate thrust. Furthermore, advanced locomotive features that significantly enhance agility—such as leaping ability, decoupled motion, and amphibious operation—are awarded additional points on a scaled system (0.25, 0.5, or 0.75), based on the extent to which these features contribute to the robot's overall manoeuvrability and effectiveness in diverse aquatic environments. This scoring system aims to provide a quantifiable and standardised method for comparing robotic agility, while also allowing for the nuanced evaluation of sophisticated motion capabilities that go beyond basic DOFs.

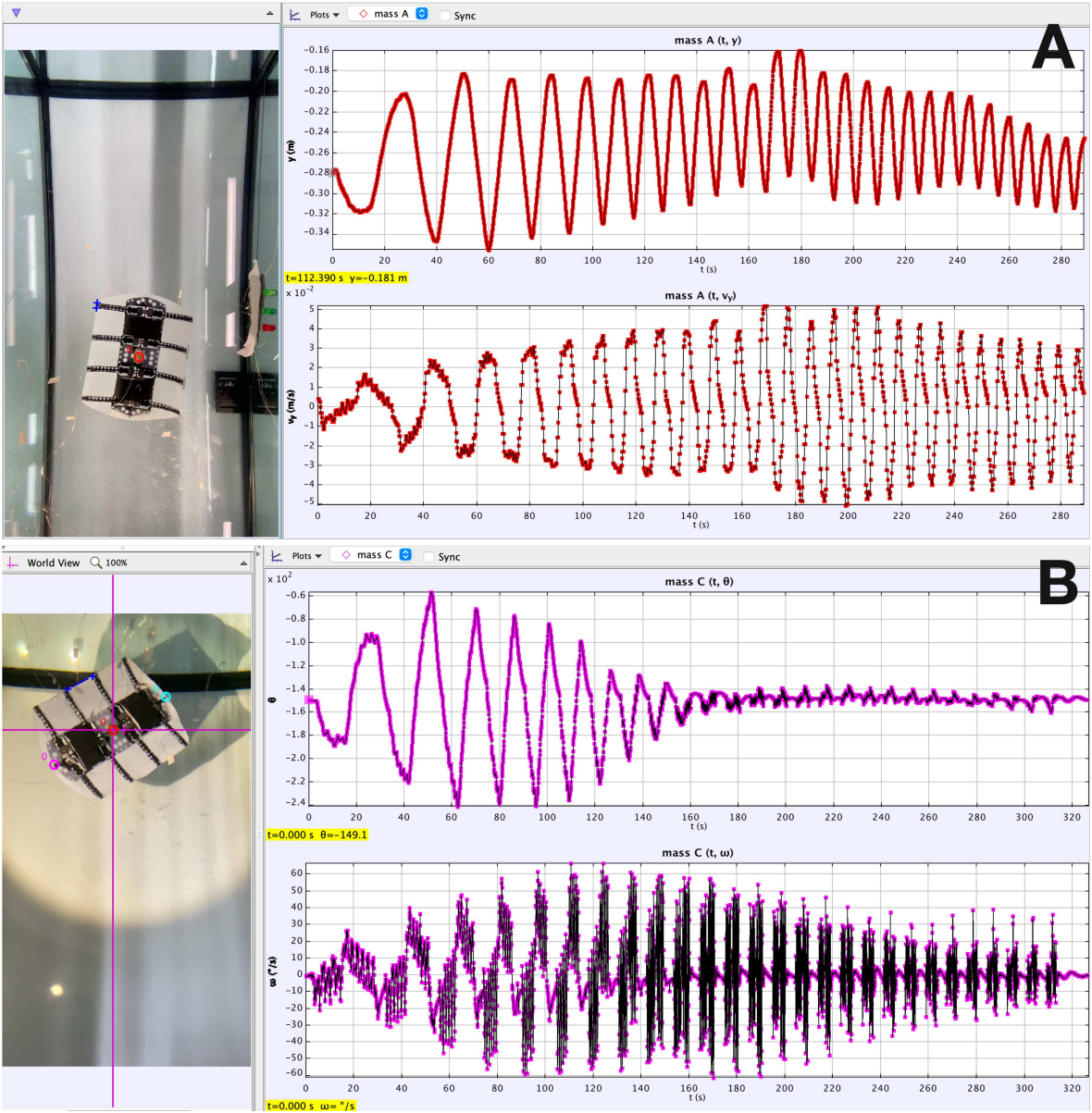

**Figure 10. Video analysis of the CuttleBot swimming back-and-forth and rotating using Physlets Tracker**

**(A)** Straight Swimming (Surge): Displacement $y$ and velocity $v_y$ over the duration of one experiment, where the wavelength = 150 mm and the frequency is varied from 0.5 to 14 Hz. The point tracked is marked in red. **(B)** Turning (Yaw): Relative angle $\Theta$ between the centre (red) and one end of the robot (purple), and angular velocity $\omega$.



\end{document}